\documentclass[journal]{IEEEtai}

\usepackage[colorlinks,urlcolor=blue,linkcolor=blue,citecolor=blue]{hyperref}
\usepackage{array}
\usepackage[table]{xcolor}
\usepackage{pifont}
\usepackage[caption=false,font=normalsize,labelfont=sf,textfont=sf]{subfig}
\usepackage{stfloats}
\usepackage{verbatim}
\let\labelindent\relax
\usepackage{enumitem}
\setlist[itemize]{leftmargin=*}
\usepackage{cite}
\usepackage{xcolor}
\usepackage{diagbox}
\usepackage{amsmath,amssymb,amsfonts}
\usepackage{graphicx}
\usepackage{textcomp}
\usepackage{url}
\usepackage{algorithm}
\usepackage{algorithmic}
\usepackage{orcidlink} %
\usepackage{titlesec}

\newcommand{\observation}{o}
\newcommand{\latent}{z}
\newcommand{\window}{w}
\newcommand{\history}{h}
\newcommand{\context}{\mathcal{M}}
\newcommand{\cS}{\mathcal{S}}
\newcommand{\cA}{\mathcal{A}}
\newcommand{\cO}{\mathcal{\MakeUppercase{\observation}}}
\newcommand{\transition}{\mathcal{T}}

\newcommand{\bR}{\mathbb{R}}

\newcommand{\horizon}{T}
\newcommand{\agentNum}{i}
\newcommand{\magnet}{l}
\newcommand{\sceneNum}{n}

\newcommand{\CandidateNum}{k}
\newcommand{\IterNum}{j}
\newcommand{\dataset}{\mathcal{D}}
\newcommand{\trajectory}{\tau}
\newcommand{\mdp}{\mathcal{M}}
\newcommand{\expect}{\mathbb{E}}

\DeclareMathOperator*{\argmax}{arg\,max}

\newcommand{\md}{{TrafficGamer}}

\makeatletter
\numberwithin{equation}{section}
\def\singlespace{\def\baselinestretch{1}\@normalsize}

\renewcommand{\baselinestretch}{1.1}

\newtheorem{definition}{Definition}

\usepackage{rotating}
\usepackage{bbm} 
\usepackage{chngcntr} 
\begin{document}

\markboth{Journal of IEEE Transactions on Artificial Intelligence}%
{TrafficGamer}

\title{TrafficGamer: Reliable and Flexible Traffic Simulation for Safety-Critical Scenarios with Game-Theoretic Oracles} 

\author{Guanren Qiao~\orcidlink{0009-0002-1351-5698}, Guorui Quan, Jiawei Yu, Shujun Jia, and Guiliang Liu~\orcidlink{0000-0001-7464-523X}, \IEEEmembership{Member, IEEE}
\thanks{This work is supported in part by Shenzhen Fundamental Research Program under grant JCYJ20230807114202005, Shenzhen Science and Technology Program under grant KJZD20240903104008012, Shenzhen Science and Technology Program under grant ZDCY20250901113000001, CUHK-CUHK(SZ)-GDSTC Joint Collaboration Fund No. 2025A0505000053,  and GuangDong Key Laboratory of
Big Data Computing (2021B1212040002).}
\thanks{Guanren Qiao is with the Chinese University of Hong Kong, Shenzhen, Guangdong, 518172 China (e-mail: guanrenqiao1@link.cuhk.edu.cn).}
\thanks{Guorui Quan is with the University of Manchester, Manchester, M13 9PL England (e-mail: 
guorui.quan@postgrad.manchester.ac.uk).}
\thanks{Jiawei Yu is with the Chinese University of Hong Kong, Shenzhen, Guangdong, 518172 China (e-mail: jiaweiyu1@link.cuhk.edu.cn).}
\thanks{Shujun Jia (corresponding author) is with Shenyang MXNavi Co., Ltd, Shenyang, 110169 China (e-mail: 
jiashujun@mxnavi.com).}
\thanks{Guiliang Liu (corresponding author) is with the Chinese University of Hong Kong, Shenzhen, Guangdong, 518172 China (e-mail: liuguiliang@cuhk.edu.cn).}
}

\maketitle

\begin{abstract}
While modern Autonomous Vehicle (AV) systems can develop reliable driving policies under regular traffic conditions, they frequently struggle with safety-critical traffic scenarios. This difficulty primarily arises from the rarity of such scenarios in driving datasets and the complexities associated with predictive modeling of multiple vehicles. Effectively simulating safety-critical traffic situations is therefore a crucial challenge. 
In this paper, we introduce TrafficGamer, which facilitates game-theoretic traffic simulation by viewing common road driving as a multi-agent game. When we evaluate the empirical performance across various real-world datasets, TrafficGamer ensures both the \textit{fidelity}, \textit{exploitability}, and \textit{diversity} of the simulated scenarios, guaranteeing that they not only statically align with real-world traffic distribution but also efficiently capture equilibria for representing safety-critical scenarios involving multiple agents compared with other methods. Additionally, the results demonstrate that TrafficGamer provides highly flexible simulations across various contexts. Specifically, we demonstrate that the generated scenarios can dynamically adapt to equilibria of varying tightness by configuring risk-sensitive constraints during optimization. We have provided a demo webpage at \url{https://qiaoguanren.github.io/trafficgamer-demo/}.
\end{abstract}

\begin{IEEEImpStatement}
This paper addresses a critical challenge in AV development: safely handling rare but high-stakes traffic scenarios, such as sudden cut-ins, aggressive merges, or emergency braking. These events are underrepresented in real-world datasets and difficult to simulate convincingly with existing tools, which often lack behavioral realism or scenario diversity. To bridge this gap, we introduce TrafficGamer, a learning-based framework that models traffic as a multi-agent game, where vehicles strategically interact in realistic and flexible ways. Engineers can leverage this tool to model diverse driving behavior and multi-vehicle interaction, generate safety-critical edge cases, control risk levels, and systematically evaluate AV performance under challenging conditions, without relying solely on costly real-world data collection.
\end{IEEEImpStatement}


\section{Introduction}
\label{sec:introduction}
\IEEEPARstart{A}{s} a cutting-edge technology, autonomous driving is a critical component of future transportation systems. The development of Autonomous Vehicles (AV) requires extensive testing and calibration of their control systems. Given the significant risks involved in conducting these tasks on real roads, the industry commonly relies on traffic simulation systems to ensure the safe development of AVs\cite{Li2024simulatorsurvey}.

Major advancements made in traffic simulation have primarily focused on improving fidelity by replicating observed vehicle trajectories. While these methods can guarantee the expected accuracy of reproducing real-world traffic flows, they often fail to capture rare events that occur at the long-tail end of the data distribution.
Conversely, a significant challenge in the design of modern AVs is their difficulty in managing these safety-critical "tail-end" events, rather than the more commonly observed traffic scenarios. This aspect underscores the importance of
actively simulating safety-critical but infrequent events, which are crucial for thoroughly testing the reliability and robustness of AV control systems.

\begin{figure*}[!t]
\centerline{\includegraphics[width=2\columnwidth]{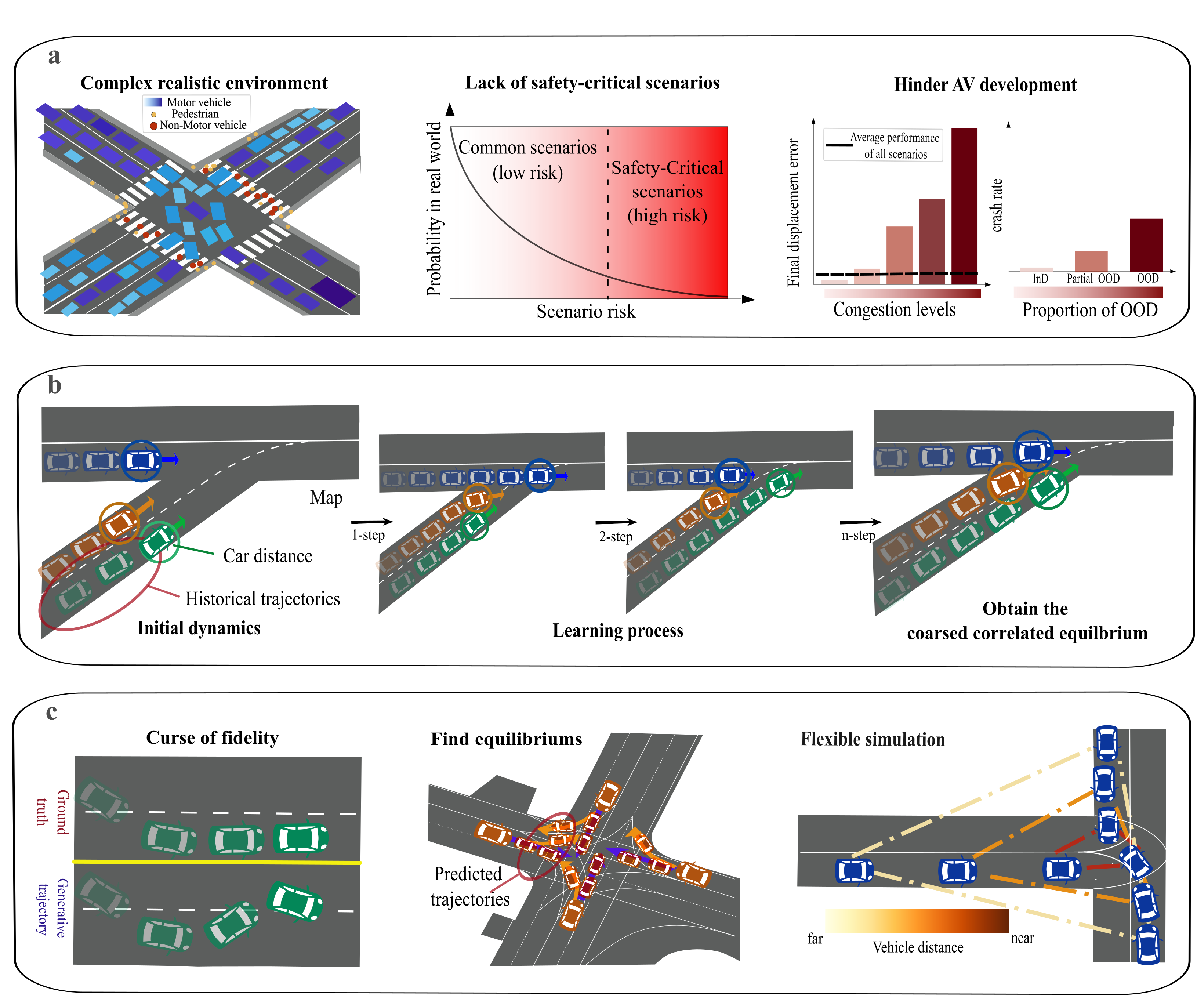}}
\caption{{\bf Motivation of crafting flexible and reliable driving scenarios with game-theoretic oracles.} Complex autonomous driving environments exist in reality, but the probability of these scenarios occurring is low. We lack data on such safety-critical instances to train a robust AV system. We find that, with an increase in congestion level and Out-of-Distribution (OOD) data, the performance of the AV system decreases. This greatly hinders the development of AV systems. TrafficGamer effectively leverages spatiotemporal information to generate safety-critical scenarios by capturing CCE. Key challenges include the curse of fidelity in trajectory prediction, finding equilibrium during CCE resolution, and achieving flexible equilibrium to control CCE intensity.}\label{fig: overview}
\end{figure*}

Beyond replicating realistic traffic scenarios, some recent studies\cite{zhou2023qcnet, wu2024smart, huang2024versatile, yan2023learning} have specifically focused on modeling vehicle behaviors or motion prediction. However, as shown in Fig.~\ref{fig: overview}a, safety-critical traffic scenarios involve complex systems with multiple agents. These high-risk events are underrepresented in real-world demonstration data, resulting in higher predictive errors and crash rates in the imitation models. Such discrepancies impact the effectiveness of data imitation and industrial experience.
More importantly, the agents' behaviors in these systems are intricately interconnected and interdependent, and the reactions of one agent are largely dependent on the movements of others. Moreover, these reactions are shaped by the agents' specific objectives and shared safety concerns; referring only to the log reply is insufficient. There is a lack of mechanisms that can effectively model the strategic behaviors of AV agents and actively formulate safety-critical events conditional on different environments. 

To this end, a critical step toward achieving accurate strategic traffic simulations is to develop a game-theoretic traffic simulation algorithm designed to model the complex interactions among multiple agents in traffic scenarios. In autonomous driving environments, vehicles and traffic participants exhibit both cooperative behaviors (e.g., collision avoidance) and implicit competition (e.g., right-of-way contention at intersections)\cite{zheng2025bilevelMARL}. The General-Sum Game framework effectively models these interactions, characterized by non-zero-sum and asymmetric payoff structures. Unlike the NE\cite{Hu2003NQL}, which enforces the independence of each agent during optimization, a CCE permits interdependencies among the agents' policies, allowing each agent's strategy to be informed by the strategies of others as shown in Fig. \ref{fig: overview}b. Table \ref{tab:equilibrium_comparison} summarizes the advantages and disadvantages of various equilibrium concepts and explains why we choose to solve for the CCE\cite{zhou2024game}.

\begin{table*}[htbp]
\centering
\caption{Comparison of Different Equilibrium Types for Autonomous Driving Systems}
\small
\begin{tabular}{p{0.22\textwidth} p{0.62\textwidth} p{0.08\textwidth}}
\hline
\textbf{Equilibrium Type} & \textbf{Advantages/Disadvantages} & \textbf{Suitability} \\
\hline

\textbf{Nash Equilibrium (NE)} 
& Solving Nash equilibrium is PPAD-complete in general games, meaning it is computationally difficult for real-world traffic game systems. Nash conditions are overly restrictive; it assumes players have complete knowledge of others' behaviors, which is hard to achieve in real-world multi-vehicle autonomous driving scenarios. Each vehicle cannot fully understand the state or future actions of other vehicles.
& \textcolor{red}{\ding{55}} \\

\hline

\textbf{Correlated Equilibrium (CE)} 
& CE coordinates players' actions via an exact public signal, where each player selects their strategy based on this signal. However, autonomous driving is a dynamic environment, and it is challenging to find a fixed, precise public signal to coordinate vehicles.
& \textcolor{red}{\ding{55}} \\

\hline

\textbf{Coarse Correlated Equilibrium (CCE)} 
& CCE allows vehicles to make decisions based on rough signals without needing complete knowledge of other vehicles' states or behaviors. Vehicles can adjust their actions based on simple signals from other vehicles or the traffic system (e.g., obstacles ahead, need to slow down, etc.). This makes CCE well-suited for environments with incomplete information, such as the interaction between vehicles in dynamic traffic flow. Moreover, CCE has the lowest computational complexity.
& \textcolor{green}{\checkmark} \\

\hline
\end{tabular}
\label{tab:equilibrium_comparison}
\end{table*}

To ensure the effectiveness of these simulated safety-critical scenarios, it is essential to resolve the following challenges shown in Fig.~\ref{fig: overview}c:
(1) {\it Flexible Simulation}: How can we construct a variety of controllable and interpretable interactive safety-critical scenarios? (2) {\it Distributional Fidelity}: How should generate vehicle trajectories closely replicate the behaviors of human drivers to provide realistic driving scenarios? A fundamental trade-off exists between realism and flexibility across both paradigms, making it difficult to achieve both concurrently in various scenarios\cite{rowe2024ctrlsim}.

Inspired by the paradigm of learning foundations models, we propose a two-phase framework for scenario generation: pre-training and fine-tuning. Specifically, we first conduct open-loop autoregressive imitation learning to pre-train based on the trajectory prediction task to capture trajectory-level realism and multimodality inherent in human driving behaviors. Subsequently, we perform closed-loop reinforcement learning (RL)\cite{Sutton1998RL, chen2025riftclosedlooprlfinetuning} combined with game theory fine-tuning in some physics-based traffic scenarios to discover a variety of safety-critical scenarios. The contributions of this
article are summarized as follows:
\begin{enumerate}
    \item The AV transportation's complexity arises not only from the diversity of road and traffic conditions but also from the increasing interactions between AVs. Many AV systems rely on existing datasets for testing; However, these datasets often fail to capture safety-critical events that, while rare in real-world scenarios, present substantial risks. To address this problem, we propose a novel framework TrafficGamer, which incorporates game-theoretic principles and proposes a new paradigm: \textbf{Equilibrium-driven Generation}. We treat multi-vehicle interaction scenarios as a general multi-agent game, redefining the scenario generation problem as \textit{finding and adjusting the equilibrium points of a multi-agent game}. 

    \item We introduce the controllable multi-agent CCE-Solver algorithm, designed to efficiently model vehicle policies that approximate the CCE. The vehicle policies derived from our model can be mathematically characterized by a CCE. To generate a range of safety-critical scenarios, we incorporate two key mechanisms: 1) Vehicle constraints, which are integrated through Lagrangian optimization to precisely regulate competitive intensity, and 2) Risk sensitivity, implemented using Conditional Value-at-Risk (CVaR), allowing for flexible control over the conservatism or aggressiveness of driving behaviors. This mechanism is controllable, enabling the discovery of safety-critical scenarios across different traffic densities.

    \item We validate the proposed approach across four comprehensive dimensions: fidelity, exploitability, flexibility, and generalization. Our experiments span two large-scale public datasets, providing a broader and more challenging evaluation setting.
\end{enumerate}

This article is organized as follows. Related work is given in Section~\ref{sec:related_work}. Section~\ref{sec:Preliminaries} illustrates the problem formulation. Section~\ref{sec: trafficgamer} describes our method. Section~\ref{sec: experimental results} demonstrates the performance of  \md\;under different datasets. Section ~\ref{sec: conclusion} concludes the paper and outlines future work.

\begin{figure*}[htbp]
\centerline{\includegraphics[width=2\columnwidth]{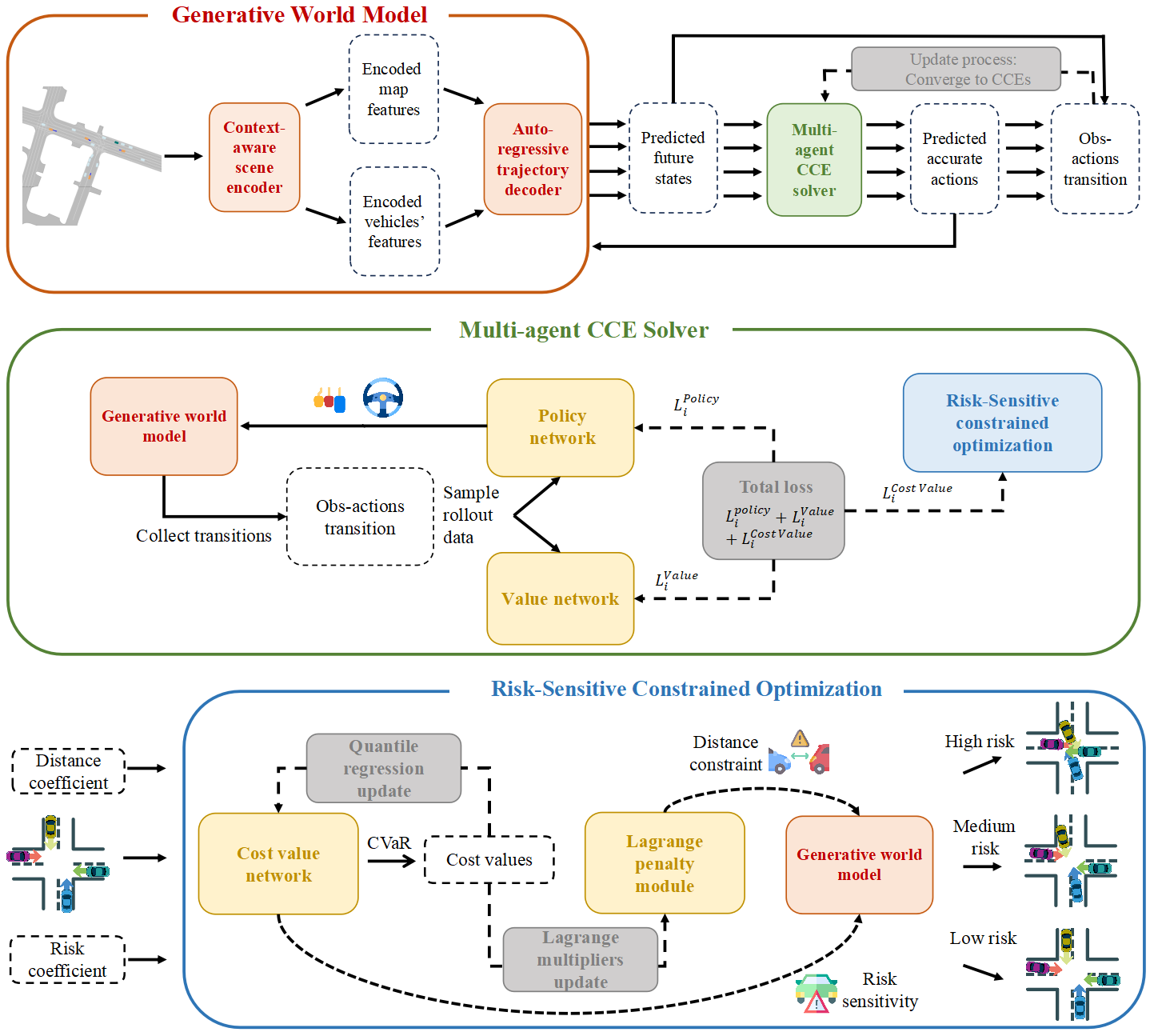}}
\caption{{\bf The structure of TrafficGamer}. The TrafficGamer framework consists of two parts: pre-training and fine-tuning. Initially, we utilize the training dataset to train a generative world model. Subsequently, MARL algorithms are employed to fine-tune this world model. We treat the world model as the environment providing observations (Obs), with throttle and brake actions for vehicles, and design reward and cost functions to establish a Decentralized Partially Observable Markov Decision Process. Our Multi-agent CCE-Solver, combined with {\it CCE-V-learning} and {\it magnet mirror descent}, is utilized to learn optimal policies for each agent. This process controls and captures the CCE during the learning phase. We adjust {\it distance constraint} and {\it risk coefficient} to control the intensity of the CCE, thereby influencing the competitiveness of vehicles in various scenarios. By extension, we combine a {\it Lagrangian-based optimization} algorithm, which adopts constraints on different vehicle distances to control the CCE, and a {\it risk-sensitive} algorithm, which employs CVaR with different confidence levels, to affect the vehicles' risk sensitivity.} 
\label{fig: framework}
\end{figure*}

\section{Related Work}
\label{sec:related_work}

{\bf Simulation-Based Testing for Autonomous Driving Systems.} Simulation-based testing has emerged as a dominant method for evaluating AVs, providing a cost-effective and safe alternative to real-world trials. Using virtual environments such as CARLA\cite{Dosovitskiy2017carla} and MetaDrive\cite{Li2023metadrive}, researchers can assess AV performance under various driving conditions, from routine traffic to rare, hazardous scenarios that are impractical or unsafe to recreate physically. Within these simulation systems, the commonly studied goals
include {\it traffic flow simulation} (e.g., SUMO\cite{Lopez2018SUMO} and CityFlow\cite{Zhang2019cityflow}), {\it sensory data simulation} (e.g., Airsim\cite{airsim2017fsr} and Unisim\cite{yang2023unisim}), {\it driving policy simulation} (e.g., Metadrive, Waymax\cite{Gulino2023waymax}, GPUDrive\cite{kazemkhani2025gpudrive} and CarDreamer\cite{Gao2024CarDreamer}), {\it vehicle dynamics simulation} (e.g., Carsim and Matlab) and {\it multi-task simulation} (e.g., CARLA).

{\bf Safety-Critical Scenario Generation.} Some prior studies have specifically focused on modeling safety-critical events.
These event generating strategies can be categorized as follows\cite{Ding2023Safety-Criticalsurvey}: (1) \textit{Data-driven generation} detects and reproduces safety-critical scenarios recorded in real-world datasets. For example, GameFormer\cite{Huang2023gameformer}, QCNet\cite{zhou2023qcnet}, SMART\cite{wu2024smart}, Donut\cite{knoche2025donut} and GUMP\cite{Hu2024GUMP} have modeled the distribution of traffic scenarios by maximizing the likelihood of observed vehicle trajectories.
(2) \textit{Adversarial generation} intentionally creates risky scenarios by manipulating the generation process of AV systems. 
In this setting,  Advsim\cite{Wang2021advsim} has manipulated the initial conditions of the scenario or provided the complete trajectory upfront.
 RL-based methods constructed an adversarial policy network\cite{Hanselmann2022king} to control autonomous vehicles. \cite{mei2025LLM-Attacker} has utilized multiple LLM agents to identify optimal attackers. Then, the trajectories of the attackers are optimized to generate adversarial scenarios. 
(3) \textit{Knowledge-based generation} leverages external domain knowledge to facilitate the generation of safety-critical events. 
To apply this strategy, ChatScene\cite{Zhang2024ChatSceneKS} has incorporated latent embeddings or constraint signals of traffic rules into their trajectory prediction models. Most of previous research has proposed learning-based models to replicate agent behaviors identified in real-world driving trajectory datasets. However, it is important to note that the number of high-risk scenarios in these datasets is limited. Therefore, it is necessary to consider generating a wider variety of scenarios with different risk levels to enhance the robustness of AV models.

Recently, some methods have proposed more controllable behavior simulation models by training generative models, integrating LLM-based scenario reasoning, or introducing RL approaches. Huang et al. and Wang et al. have utilized diffusion generative models to predict scene-consistent and controllable multi-agent interactions in closed-loop settings\cite{huang2024versatile, wang2024optimizing}. Chang et al.\cite{chang2023editing} can adjust the level of courtesy in the generated trajectory while ensuring that the trajectories remain realistic and human-like by learning from real driving data. Rowe et al. have trained a return-conditioned multi-agent behavior model that allows for fine-grained manipulation of agent behaviors by modifying the desired returns for the various reward components\cite{rowe2024ctrlsim}. \cite{mei2025seeking}, ChatSUMO\cite{li2024chatsumo} and Chat2scenario\cite{zhao2024chat2scenario} have utilized LLM to understand and generate different driving scenarios by incorporating high-level contextual knowledge. However, the most works mentioned so far is either limited to 2–3 agents when generating scenarios, or lacks interpretability regarding the behaviors of the agents. 

\section{Preliminaries}
\label{sec:Preliminaries}

We formulate the task of multi-vehicle motion prediction in the traffic scenarios as a {\bf Partially Observable Constrained Markov Decision Process (POCMDP).} 
$( \cS, \{\Omega_\agentNum,\cA_\agentNum, \cO_\agentNum,r_\agentNum, c_\agentNum\}_{\agentNum=1}^{\MakeUppercase{\agentNum}},\transition,\gamma,p_0 )$ where:
\begin{itemize}
    \item $\agentNum$ denotes the number of agents from 1 to $\MakeUppercase{\agentNum}$.
    \item $\Omega_\agentNum$ and $\cA_\agentNum$ denote the spaces of observations and actions for a specific agent $\agentNum$. The observations include the position, velocity, heading, and partial map features in the surrounding neighbor region of the agent $\agentNum$. The actions consist of relative heading and acceleration, as described in\cite{Gulino2023waymax}. We constrain the output actions within a reasonable range to prevent irrational driving behaviors such as sudden acceleration, sharp turns, etc. 
    \item $\cS$ denotes the state space that comprises all agents' historical trajectory information
    and map features.
    \item $\cO_\agentNum:\cS\rightarrow\Omega_\agentNum$ denotes the observation function that maps states and actions to local observation for the $\agentNum^{th}$ agent. $\boldsymbol{\cO}=\{\cO_1,\dots,\cO_{\MakeUppercase{\agentNum}}\}$ denotes the function set.
    \item $r_\agentNum(\cdot):\{\Omega_\agentNum\times\cA_\agentNum\}_{\agentNum=1}^{\MakeUppercase{\agentNum}}\rightarrow\bR$ denotes the agent-specific reward function 
    that maps actions and observations from all agents to the reward of the $\agentNum^{th}$ agent. We consider reward factors such as avoiding collisions, lane deviation, and reaching the destination, among others\cite{Li2023metadrive}.
    \item $c_\agentNum(\cdot):\{\Omega_\agentNum\times\cA_\agentNum\}_{\agentNum=1}^{\MakeUppercase{\agentNum}}\rightarrow\bR$ denotes the agent-specific cost function that maps actions and observations from all agents to the cost of $\agentNum^{th}$ agent. We adapt the vehicle distance as a constraint condition. Additionally, the expectation of cumulative constraint functions $c_\agentNum$ must not exceed the constraint threshold $\delta$.
    \item $\transition:\cS\times\cA\rightarrow\Delta^{\cS}$ \footnote{$\Delta^{\cS}$ denotes the probability simplex over the space $\cS$.}denotes the transition function.
    \item $\gamma\in[0,1]$ and $p_0\in\Delta^{\cS}$ denote the discount factor and the initial distribution, respectively.
\end{itemize}

Specifically, we design a piecewise reward function $r_1(\cdot) = -{\|d_{total}-d_{pass}\|}$
where $d_{total}$ is the distance to the endpoint, $d_{pass}$ means the distance you have passed. $r_2(\cdot)$ detects whether the vehicle will drift off the lane. $r_3(\cdot) = -c_3$ if a collision occurs. $r_4(\cdot)$ defines whether the velocity speed is higher than the maximum limit velocity. $r_\agentNum(\tau)$ is ultimately the linear combination of the reward functions $r_1$, $r_2$, $r_3$ and $r_4$. For the cost function, we design a binary classification function $c_i(\cdot) = \|{pos}_i-{pos}_j\|_2
$
where $pos$ is denoted as the position of the agents and $j$ is the index of the other agents. If $c_\agentNum(\cdot)\leq\beta$, the cost value is 1, where $\beta$ is the distance constraint. 

We have explained the symbols involved in the following content in Table \ref{tab:symbol_explanation}.


In the framework of POCMDP, each agent is assigned an individual reward function and cost function, denoted as $r_\agentNum(\cdot)$ and $c_\agentNum(\cdot)$. This aligns with a general-sum game structure\cite{Mao2023GS}, which is more complex and less explored than zero-sum and cooperative games\cite{Yang2020overview}. Despite the challenges, modeling a general-sum game better aligns with real-world driving scenarios, as human drivers often prioritize their objectives. These objects can be applied to satisfy conflicting interests among different vehicles. During the optimization process, human drivers' behaviors inevitably affect others' decisions.
Accordingly, we consider the {\bf General Sum Markov Games (GS-MGs)} under the POCMDP. For each agent $\agentNum$, the value function $V^{\boldsymbol{\pi}}_{\agentNum,t}:\cS\rightarrow\bR$
action-value function $Q^{\boldsymbol{\pi}}_{\agentNum,t}:\cS\times\cA\rightarrow\bR$ and advantage function $A^{\boldsymbol{\pi}}_{\agentNum,t}:\cS\times\cA\rightarrow\bR$ are represented by:
\begin{align}
    &V^{\boldsymbol{\pi}}_{\agentNum,0}(s)=\expect_{p_0,\transition,\boldsymbol{\pi}}\left[\sum_{t=0}^{\horizon}\gamma^tr_\agentNum(\boldsymbol{\observation}_{t}, \boldsymbol{a}_{t})|\boldsymbol{o}_{\agentNum,0}=\boldsymbol{\cO}_{\agentNum}(s_{t+1})\right]\label{obj:v-function}
\end{align}
\begin{equation}
\begin{aligned}
    Q^{\boldsymbol{\pi}}_{\agentNum,0}(s, a_\agentNum,-\boldsymbol{a}_\agentNum)=&\expect_{p_0,\transition,\boldsymbol{\pi}}\Big[\sum_{t=0}^{\horizon}\gamma^tr_\agentNum(\boldsymbol{\observation}_{t}, \boldsymbol{a}_{t})|\\&\boldsymbol{o}_{\agentNum,0}=\boldsymbol{\cO}_{\agentNum}(s_{t+1}), a_{\agentNum,0}=a_\agentNum\Big]\label{eqn:q-function}
\end{aligned}
\end{equation}
\begin{align}
    &A^{\boldsymbol{\pi}}_{\agentNum,0}(s, a_\agentNum,-\boldsymbol{a}_\agentNum)=Q^{\boldsymbol{\pi}}_{\agentNum,0}(s, a_\agentNum,-\boldsymbol{a}_\agentNum)-V^{\boldsymbol{\pi}}_{\agentNum,0}(s)\label{eqn:advantage-function}
\end{align}
where $-\boldsymbol{a}_\agentNum=\{\mathbf{1}_{\agentNum^\prime\neq\agentNum}a_\agentNum^\prime\}_{\agentNum^\prime=1}^{\MakeUppercase{\agentNum}}$\footnote{Throughout this work, the bold symbols (e.g., $\boldsymbol{a}$) indicate a vector of variables, while the non-bold ones (e.g., $a$) represent a single variable.} denotes the joint action performed by $\MakeUppercase{\agentNum}-1$ players (without $\agentNum$'th player) and $\boldsymbol{\pi}=\{\pi_\agentNum\}_{\agentNum=1}^{\MakeUppercase{\agentNum}}$ denotes the product policy.

Under a GS-MG, the goal of policy optimization is to capture a {\bf  Coarse Correlated Equilibrium}  
(Definition \ref{def:cce-gap})\cite{ToroBetancur2023CCE}. 

\begin{definition}
    \textit{($\epsilon$-approximate CCE). A General Correlated policy $\boldsymbol{\pi}$\cite{Song2022SE} is an $\epsilon$-approximate Coarse Correlated Equilibrium ($\epsilon$-CCE) if
    \begin{align}
        \max_{\agentNum\in[\MakeUppercase{\agentNum}]}\left(V^{\dagger,\boldsymbol{\pi}_{-\agentNum}}_{\agentNum,0}(s)-V^{\boldsymbol{\pi}}_{\agentNum,0}(s)\right)\leq\epsilon
    \end{align}
    where $V^{\dagger,\boldsymbol{\pi}_{-\agentNum}}_{\agentNum,0}(s)=\sup_{\pi^\prime_{\agentNum}}V^{\pi^\prime_{\agentNum},\boldsymbol{\pi}_{-\agentNum}}_{\agentNum,0}(s)$ denotes the best response for the $\agentNum^{th}$ agent against $\pi_{-\agentNum}$. Unlike traditional policy in POMDPs, $\pi$ inputs observations that include historical vehicle state information. We say $\boldsymbol{\pi}$ is an (exact) CCE if the above is satisfied with $\epsilon=0$.}\label{def:cce-gap}
\end{definition}


Unlike previous GS-MG solvers\cite{Hu2003NQL,Song2022SE,Hambly2023Games,Mao2023GS} that rely on an interactive environment, traffic simulation presents additional challenges, as our algorithm can only utilize an offline database with records of the behaviors of multiple drivers on open roads. Our algorithm aims to learn the CCE policies of multiple AVs under various constraints that accurately reflect the human drivers' behaviors in the dataset. This problem can be formulated as {\bf Offline Multi-agent Constrained Reinforcement Learning (Offline MA-CRL)}. Specifically, the problem can be summarized as follows:
\begin{definition}\label{def:Offline-MA-RL} \textit{(Offline MA-CRL in GS-MGs.)
let $\dataset_l=\{\context_\sceneNum,\trajectory_{\sceneNum,1},\dots,\trajectory_{\sceneNum,\MakeUppercase{\agentNum}}\}_{\sceneNum=1}^{\MakeUppercase{\sceneNum}}$ defines the offline dataset, where $\sceneNum=[\MakeUppercase{\sceneNum}]$ defines the number of scenarios, $\context_\sceneNum$ represents the game context in the $\sceneNum^{th}$ scenario, $\boldsymbol{c} = \{{c_0,c_1,...,c_\agentNum}\}_{\agentNum=1}^{\MakeUppercase{\agentNum}}$ represents constraints and $\trajectory_{\sceneNum,\agentNum}=\{o_{\agentNum,0},a_{\agentNum,0},\dots,o_{\agentNum,\horizon},a_{\agentNum,\horizon}\}$ denotes the trajectory of $\agentNum^{th}$ agent in the $\sceneNum^{th}$ scenario. Given $\dataset_l$, the goal of our algorithm is to learn a $\boldsymbol{\hat{\pi_{\boldsymbol{c}}}}$ that satisfies the constraints with the following properties: (1) {\it Exploitability}: $\boldsymbol{\hat{\pi_{\boldsymbol{c}}}}$ satisfies the $\epsilon$-approximate CCE in Definition~\ref{def:cce-gap}, (2) {\it Fidelity}: $\boldsymbol{\hat{\pi_{\boldsymbol{c}}}}$ must be consistent with the real driver's policies such that $D_f(\boldsymbol{\hat{\pi_{\boldsymbol{c}}}},\boldsymbol{\pi}^l)\leq\xi$ where $D_f$ and $\xi$ denote the divergence metric and a threshold, respectively, and (3) {\it Flexibility}: $\boldsymbol{\hat{\pi_{\boldsymbol{c}}}}$ satisfies the CCEs of varying intensities.}
\end{definition}

Safety-critical scenarios are common in real-world driving environments. Therefore, Definition~\ref{def:Offline-MA-RL} shows that the gap between the safety-critical scenarios generated by our method and those encountered in real-world driving situations should be minimized as much as possible.

\section{Method}
\label{sec: trafficgamer}

In this work, to solve the offline MA-CRL problem in GS-MGs under the POCMDP (Definition \ref{def:Offline-MA-RL}), our TrafficGamer model considers a model-based MA-CRL approach as shown in Fig. \ref{fig: framework} that (1) \textbf{trains a generative world model} to acquire environment features in a data-driven way, (2) \textbf{converges to CCEs} based on the predicted environment dynamics and predefined action space and (3) adjusts the level of competition for \textbf{capturing diverse degrees of CCEs}.

\subsection{Modelling Traffic Dynamics Based on Offline Data}\label{subsec:world-model}

To represent the environmental dynamics, we introduce the generative world model based on the offline datasets $D_l$.

{\bf Generative World Model.} For computation efficiency, we follow the decoder-only architecture in \cite{zhou2023qcnet} and implement the world model as an \textit{Auto-regressive Trajectory Decoder}. As a sequential prediction model, our world model maps the previous state (e.g., $s_{t-1}$) and the action of each agent (e.g., $a_{1,t},\dots, a_{\agentNum,t}$) onto the next state $s_t$.
Unlike QCNet\cite{zhou2023qcnet}, which generates trajectories for future vehicle motions over a fixed period, our world model adopts an auto-regressive approach, predicting vehicles' step-wise motion based on its previous predictions.  This approach enables the modeling of how past movements influence future decisions. 

\begin{figure}[htbp]
\centering
\includegraphics[scale=0.14]{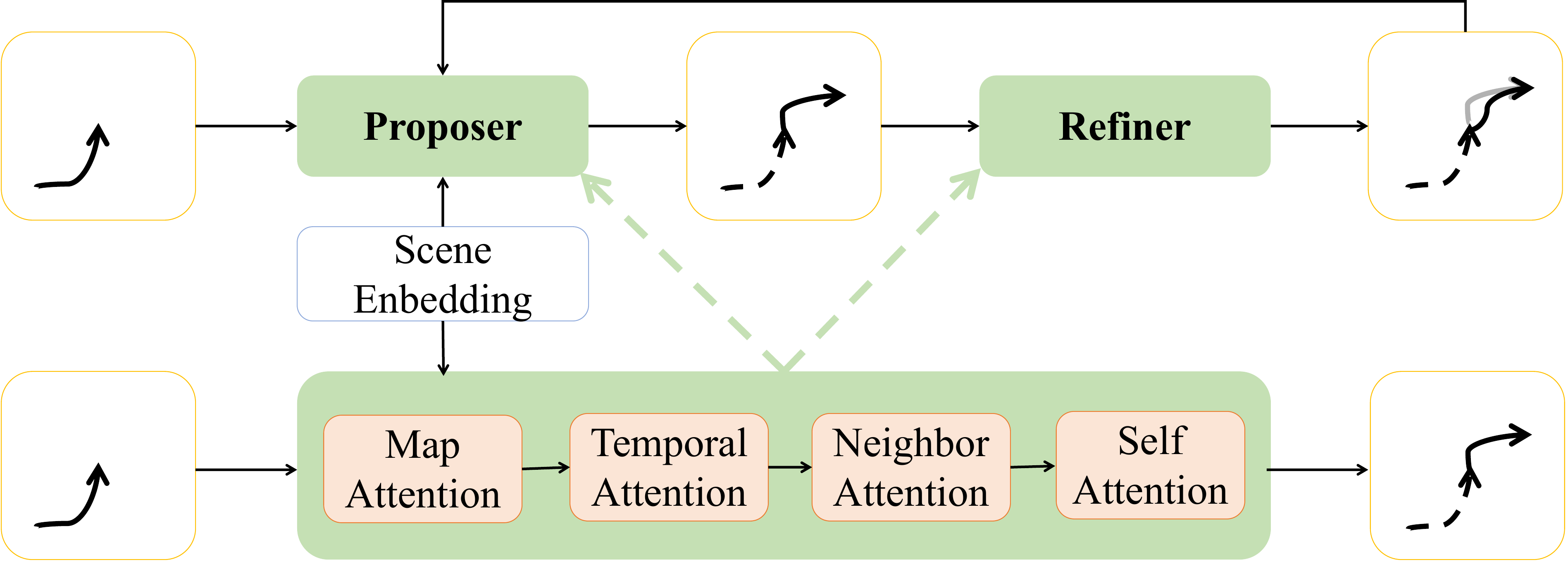}
\caption{{\bf Illustration of the decoder module.} We have some attention layers to decode accurate trajectories. (1) {\it Agent-to-Map Cross Attention} incorporates the map information $\latent^{m}=f_e(\mathcal{M})$ ($f_e$ is a map encoder) including all kinds of lane markings and types of polygons into the most recent agents' features such that e.g., $\boldsymbol{\latent}^{s}_{\agentNum,t}=\text{CrossAttn}(\trajectory_{1,t},\dots,\trajectory_{\MakeUppercase{\agentNum},t},\latent^{m})$. 
(2) {\it Agent-to-Temporal Cross Attention} that embeds the historical information into the agents' trajectories $\trajectory_{1,t+1},\dots,\trajectory_{\MakeUppercase{\agentNum},t+1}$ where $\trajectory_{1,t+1}$ consists of $a_{1,t}$ and $s_{1,t}$, e.g.,  $\latent^{\MakeUppercase{\window}}_{\agentNum,t}=\text{CrossAttn}(\{\latent^{s}_{\agentNum,t-1},\dots,\latent^{s}_{\agentNum,t-\MakeUppercase{\window}}\},\latent^{s}_{\agentNum,t})$, where $H$ indicates the length of encoding historical steps. (3) {\it Agent-to-Neighbor Cross Attention} that embeds the spatial-temporal features of the surrounding agents such that $\latent^{N}_{\agentNum,t}=\text{CrossAttn}(\latent^{\MakeUppercase{\window}}_{\agentNum,t},\{\latent^{\MakeUppercase{\window}}_{j,t}\}_{j\in N_i})$ and (4) {\it Self-Agent Attention} considers the agent's entire predicted trajectory at each time step, capturing dependencies between distant elements, e.g., $\latent^{A}_{\agentNum,t} = \text{SelfAttn}(\{\latent^{N}_{\agentNum,t}\}^T_{t=0})$. 
The final state $s_{t+1}$ is represented by concatenating $s_{t}=\{\boldsymbol{\latent}^{A}_{\agentNum,\window}\}_{\window=t-\MakeUppercase{\window},\agentNum=0}^{t,\MakeUppercase{\agentNum}}$ with the predicted $\boldsymbol{\latent}^{A}_{1,t+1},\dots,\boldsymbol{\latent}^{A}_{\MakeUppercase{\agentNum},t+1}$.}
\label{fig: component}
\end{figure}

{\bf Action Predictor.} Under the context of POCMDP, when $t=0$, $s_{0}$ captures the static game map $\context$ and initial features of all agents. When ${t>0}, s_{t}=\{\boldsymbol{\latent}^{s}_{\agentNum,w}\}_{w=t-\MakeUppercase{\window},\agentNum=0}^{t,\MakeUppercase{\agentNum}}$ captures the spatial-temporal information of all agents under the game map $\context$ in the previous $t-1$ time steps, and $a_{1,t},\dots, a_{\agentNum,t}$ denotes the acceleration and heading of agents in the current time step. Under this setting, our decoder is implemented by (1) {\it Agent-to-Map Cross Attention}, (2) {\it Agent-to-Temporal Cross Attention}, (3) {\it Agent-to-Neighbor Cross Attention} and (4) {\it Self-Agent Attention}, which incorporates map information, historical information, the spatial-temporal features of the surrounding agents, and the features of the agent itself into the temporal dimension thereby mapping $s_{t}$ and $\boldsymbol{a}_{1,\dots,\MakeUppercase{\agentNum},t}$ to $s_{t+1}$. 
The decoder predicts actions based on state $s_{t}=\{\boldsymbol{\latent}^{s}_{\agentNum,\window}\}_{\window=t-\MakeUppercase{\window},\agentNum=0}^{t,\MakeUppercase{\agentNum}}$ to satisfy latent traffic rules in realistic driving scenarios. The decoder consists of two modules: the proposer and the refiner. The proposer is used to predict the next sub-trajectory segment, while the refiner performs finer corrections on the predicted trajectory. For the specific architecture, please refer to Fig \ref{fig: component}.
We sample the $i$-th agent’s future trajectory as a weighted mixture of Laplace distributions by following\cite{zhou2023qcnet}:
\begin{align}
    \pi^\magnet_\agentNum(\hat{\trajectory})=\prod\limits_{t=1}^\mathcal{T}\sum\limits_{\CandidateNum=1}^{\MakeUppercase{\CandidateNum}}\omega^{\CandidateNum}_{\agentNum}p\left(a_{\agentNum,t}\mid\mu_{i,t}^\CandidateNum,\vartheta_{\agentNum,t}^\CandidateNum\right)\label{eqn:imitate-policy}
\end{align}
where (1) $\omega_\agentNum^\CandidateNum$ denotes a learnable weight coefficient and (2) $\mu_{i,t}^\CandidateNum$ and $\vartheta_{\agentNum,t}^\CandidateNum$ characterize the mean position and the level of uncertainty, respectively, of the $i$-th agent at the time step $t$. We choose ${\max \limits_{\CandidateNum}} \enspace\omega_\agentNum^\CandidateNum p(a_t|\mu_{i,t}^{\CandidateNum},\vartheta_{\agentNum,t}^{\CandidateNum})$ as the $\pi_i^l$. 

{\bf Observation Model.} For each agent, our observation model maps $s_t$ and $a_{\agentNum,t}$ onto agent-specific observations $\{\boldsymbol{o}_{\agentNum,t}\}_\agentNum^I$. The observation model is implemented by $o_{\agentNum,t}=f^{\text{MLP}}_{\agentNum}(z^{A}_{\agentNum,t})$, where MLP is the Multilayer Perception. Additionally, this module considers the historical actions and observation of all agents such that $\boldsymbol{\history}=\{(\observation_{\agentNum,\iota}, a_{\agentNum,\iota})\}_{\iota=0,\agentNum=1}^{\window,\MakeUppercase{\agentNum}}$.

\subsection{Learning CCEs in the General Sum Markov Games}
As our environment is structured as a GS-MG with rewards and costs specific to each agent,  we are inspired by\cite{Mao2023GS} to consider a decentralized update of each agent's policy where we fix the rest $\MakeUppercase{\agentNum}-1$ agents' policy $\boldsymbol{\pi}_{-i}$ and train policy $\pi_\agentNum$ to get the best response of agent $i$. In this paper, we updated $\pi_\agentNum$ by iteratively optimizing the following objective in the $j^{th}$ iteration
\begin{equation}
\begin{aligned}
\pi^{\IterNum}_{\agentNum} = \argmax \,
    & \expect_{\pi_{\agentNum},\mu_0}\left[ V^{\pi_\agentNum,\boldsymbol{\pi}_{-i},j-1}_{\agentNum,t}(s) \right] \\
    & - \eta_1 \mathcal{B}_{\psi}(\pi_{\agentNum}, \pi^\magnet_\agentNum) 
     - \frac{1}{\eta_2} \mathcal{B}_{\psi}(\pi_{\agentNum}, \pi^{\IterNum-1}_{\agentNum})
\end{aligned}
\label{obj:marl-mmd}
\end{equation}
where $\pi^\magnet_\agentNum$ denotes the imitation policy learned by the world model (Equation~\ref{eqn:imitate-policy}), and the $\pi^{\IterNum-1}_{\agentNum}$ denotes the policy learned from the previous iteration. We follow the multi-player CCE-V-Learning algorithm\cite{Song2022SE} to update $V^{\pi_\agentNum,\boldsymbol{\pi}_{-i},j-1}_{\agentNum,t}$. for learning CCE in GS-MGs. This objective also contains one key component that can efficiently facilitate convergence to a CCE by utilizing:

{\bf Magnetic Mirror Descent (MMD).} We follow\cite{Sokota2023MMD} and incorporate the Bregman divergence $\mathcal{B}_{\psi}(\cdot,\cdot)$ with respect to the mirror map $\psi$ such that $\mathcal{B}_{\psi}(x,y)=\psi(x)-\psi(y)-\langle\nabla\psi(y),x-y\rangle$ into the objective with convergence guarantees. Recent studies\cite{li2024CMD,Sokota2023MMD} have confirmed that the mirror descent approaches can solve different kinds of games in multi-player settings.
 To derive a more intuitive objective, we implement the mirror map as the negative entropy such that $\psi(x)=\sum p(x)\log p(x)$, and the objective~(\ref{obj:marl-kl-mmd}) becomes:
\begin{equation}
\begin{aligned}
    \pi^{\IterNum}_{\agentNum,t}=\argmax&\expect_{\pi_{\agentNum},\mu_0}[V^{\pi_\agentNum,\boldsymbol{\pi}_{-i},j-1}_{\agentNum,t}(s)]\\
    &-\eta_1\mathcal{D}_{KL}(\pi_{\agentNum}\|\pi^\magnet_{\agentNum})-\frac{1}{\eta_2}\mathcal{D}_{KL}(\pi_{\agentNum}\|\pi^{\IterNum-1}_{\agentNum})\label{obj:marl-kl-mmd}
\end{aligned}
\end{equation}
where $\mathcal{D}_{KL}$ is the KL-divergence of two variables. Intuitively, by punishing the distance between current policy $\pi_{\agentNum}$ and imitation policy $\pi^l_{\agentNum}$, this objective ensures the fidelity in the offline MA-CRL problem (Definition~\ref{def:Offline-MA-RL}). By constraining the scale of updates between current policy $\pi_{\agentNum}$ and previous policy $\pi^{\IterNum-1}_{\agentNum}$, the training process becomes more stable.

However, both the transition function $\transition$ and policy of other players $\boldsymbol{\pi}_{-\agentNum,t}$ are not subject to optimization in the objective~(\ref{obj:marl-kl-mmd}), and thus we consider the discounted causal entropy $\sum_{t=0}^{\horizon}\gamma^t\mathcal{H}[\pi(a_{i,t}|\observation_{\agentNum,t},\boldsymbol{h})]$. The agent-specific observation $\{\boldsymbol{o}_{\agentNum,t}\}_\agentNum^I$ of each agent is implemented by $o_{\agentNum,t}=f^{\text{MLP}}_{\agentNum}(z^{s}_{\agentNum,t})$, where MLP is the Multilayer Perception. Furthermore, the historical actions and observation of all agents are denoted as $\boldsymbol{\history}=\{(\observation_{\agentNum,\iota}, a_{\agentNum,\iota})\}_{\iota=0,\agentNum=1}^{\window,\MakeUppercase{\agentNum}}$. Similarly, instead of utilizing the computationally intractable trajectory-level KL-divergence $\mathcal{D}_{KL}(\pi_{\agentNum}\|\pi^{\IterNum-1}_{\agentNum})$, we consider the time-wise causal KL-divergence $\sum_{t=0}^{\horizon}\gamma^t\mathcal{D}_{KL}[\pi_{\agentNum,t}(\cdot)\|\pi^{\IterNum-1}_{\agentNum,t}(\cdot)]$ where we write $\pi_{\agentNum,t}(\cdot)$ instead of $\pi(a_{i,t}|\observation_{\agentNum,t},\boldsymbol{h})$ for brevity and by substituting it into the objective~(\ref{obj:marl-kl-mmd}).

Since $\mathcal{D}_{kl}(x,y)=\mathcal{H}(x,y)-\mathcal{H}(x)$, objective~(\ref{obj:marl-kl-mmd}) can be further derived as:
\begin{align}
\max\expect
    \Big[\sum_{t=0}^{\horizon}\gamma^t\Big(&r^*_{\agentNum}(\boldsymbol{\observation}_{t}, \boldsymbol{a}_{t})+\eta\mathcal{H}[\pi_{\agentNum,t}(\cdot)]\Big)\Big]\label{obj:marl-pg}
\end{align}
where for brevity, we denote $\eta=\frac{1+\eta_1\eta_2}{\eta_2}$, $\pi_{\agentNum,t}(a_{i,t}|\observation_{\agentNum,t})$ as $\pi_{\agentNum,t}(\cdot)$ and $r^*_{\agentNum}(\boldsymbol{\observation}_{t}, \boldsymbol{a}_{t}) = r_{\agentNum}(\boldsymbol{\observation}_{t}, \boldsymbol{a}_{t})+\expect_{\pi_{\agentNum,t}}[\log(\pi^l_{\agentNum,t})^{\eta_1}(\pi^{\IterNum-1}_{\agentNum,t})^{\frac{1}{\eta_2}}]$. We can treat the second entropy term as a constant. This objective also maximizes the reward of agent $\agentNum$, which aligns well with the RL paradigm.

\subsection{Controlling the Levels of Adversity among Heterogeneous Agents}
To effectively simulate complex traffic scenarios that include various vehicle types, such as cars, buses, and trucks, and diverse driving styles, such as aggressive and conservative driving, it is crucial to tailor the behavior of each agent to regulate the level of competition within the scenarios created. We can more thoroughly assess the system's robustness by subjecting the AV control system to these various scenarios. This comprehensive evaluation helps in developing trustworthy AV vehicles capable of performing reliably in realistic traffic conditions.
To accurately represent the diverse levels of scenarios characterized by different CCEs, we incorporate constrained and risk-sensitive policy optimization into the multi-agent traffic simulation system.

{\bf Constrained Traffic Simulation.} To dynamically adjust the intensity of CCEs, we request the agents to impose varying levels of driving constraints on the AV agents, thereby modulating the severity and nature of the driving conditions. Specifically, we expand the objective~(\ref{obj:marl-pg}) by formulating the trade-off between rewards and costs under a constrained policy optimization objective:
\begin{equation}
\begin{aligned}
\arg\max_{\pi}&\mathbb{E}_{\pi,\transition,\mu_{0}}\Big[\sum_{t=0}^{T}\gamma^{t}\Big(r^*_\agentNum(\boldsymbol{\observation}_{t}, \boldsymbol{a}_{t})+\eta\mathcal{H}[\pi_{\agentNum,t}(\cdot)]\Big)\Big] \\&
~\mathrm{s.t.}~ \expect\Big[\sum_{t=0}^{T}\gamma^{t}c_\agentNum(\observation_{\agentNum,t}, a_{\agentNum,t})\Big]\leq\beta \label{obj: CRL}
\end{aligned}
\end{equation}
In this study, we mainly explore how the distance constraint influences the resulting CCE from our algorithm. Additionally, we can set different vehicle distance constraints to achieve varying intensities of CCE. As the distance between vehicles increases, the competitiveness among agents decreases.

{\bf Risk-sensitive Traffic Simulation.} This strategy explicitly manages the risk sensitivity of driving behaviors, thereby deriving risk-seeking or risk-averse policies for each AV agent. This approach effectively promotes either aggressive or conservative driving policies to enhance the realism and variability of the scenarios.
With the aim of guiding risk-sensitive and constraints-satisfying policies for multiple agents, inspired by\cite{xu2024uaicrl}, 
we develop a risk-sensitive constraint to capture the uncertainty induced by environmental dynamics and extend the objective~(\ref{obj: CRL}) as follows:
\begin{equation}
\begin{aligned}
\arg\max_{\pi}&\mathbb{E}_{\pi,\transition,\mu_{0}}\Big[\sum_{t=0}^{T}\gamma^{t}\Big(r^*_\agentNum(\boldsymbol{\observation}_{t}, \boldsymbol{a}_{t})+\eta\mathcal{H}[\pi_{\agentNum,t}(\cdot)]\Big)\Big] \\
&~\mathrm{s.t.}~ \rho_{\alpha}\Big[\sum_{t=0}^{T}\gamma^{t}C_\agentNum(\mathcal{O}_\agentNum(s_t),A_{\agentNum,t})\Big]\leq\beta \label{obj: RS-CRL}
\end{aligned}
\end{equation}
where $C$ is the cost variable and $\alpha$ represents confidence. To specify the risk measure, we define the corresponding risk envelope $\mathcal{U}_{\alpha}^{\pi_\agentNum} = \{\zeta_{\alpha}: \Gamma \to [0,\frac{1}{\alpha}]|\sum_{\tau_\agentNum\in\Gamma}\zeta(\tau_\agentNum)\pi_\agentNum(\tau_\agentNum) = 1\}$, characterized as a compact, convex, and bounded set. This envelope guides the risk measure, which is induced by a distorted probability distribution for each agent $p^\zeta = \zeta \cdot p$. For example, the CVaR can be defined as $\rho_{\alpha}^{\pi_\agentNum}[\sum_{t=0}^{T}\gamma^{t}c_{\agentNum,t}]=\sup_{\zeta_{\alpha}\in\mathcal{U}_{\alpha}^{\pi_\agentNum}}\mathbb{E}_{\tau_\agentNum\sim p^{\pi_\agentNum}}[\zeta_{\alpha}(\tau_\agentNum)\sum_{t=0}^{T}\gamma^{t}c_{\agentNum,t}]$. 

Constructing the distorted probability-based risk measure relies on the estimated distribution of discounted cumulative costs. 
To estimate this distribution, we define the variable of discounted cumulative costs as $Z^c(\observation_{\agentNum,t})=\sum_{\iota=0}^{T-t}\gamma^{\iota}C_{\agentNum,\iota}|\mathcal{O}_\agentNum(s_0)=\observation_{\agentNum,t}$. 
During fine-tuning, the stochastic POCMDP process can be captured by the distributional Bellman equation\cite{dabney2018distributional}:
\begin{equation}
\begin{aligned}
    &Z^c(\observation_{\agentNum,t}):\overset{\Delta}{\operatorname*{=}}C_\agentNum(\observation_{\agentNum,t},A_{\agentNum,t})+\gamma Z^c(\cO_{\agentNum}(s_{t+1}))\\&~\text{where}~
    s_{t+1}\sim\transition(\cdot|s_t,\boldsymbol{a}_{t})~\text{and}~ \boldsymbol{a}_{t}\sim\boldsymbol{\pi}_{t}(\cdot|\boldsymbol{o}_{t},\boldsymbol{h})\label{obj: dbe}
\end{aligned}
\end{equation}
We parameterize the distribution with $N$ supporting quantiles and update these function via quantile regression, which acts as an asymmetric squared loss in an interval [$-\kappa,\kappa$] around zero:
\begin{equation}
\begin{aligned}
    &\rho_{\tau_q}^{\kappa}(u)=|\tau_q-\delta_{\{u<0\}}|\mathcal{L}_{\kappa}(u)\\&~\text{where}~\mathcal{L}_\kappa(u)=\begin{cases}\frac{1}{2}u^2,&\quad\mathrm{if} |u|\leq\kappa\\\kappa(|u|-\frac{1}{2}\kappa),&\quad\mathrm{otherwise}\end{cases}
\end{aligned}
\end{equation}
$\tau_q$ is the quantile, $\delta$ denotes a Dirac and $\mathcal{L}_{\kappa}(u)$ is a Huber loss.
Under these formulations, the risk-sensitive advantage function $A_{\agentNum, t}^{\boldsymbol{\pi},c}$ can be computed with 1-step TD updates
such that $A_{\agentNum, t}^{\boldsymbol{\pi},c}=c_{{\agentNum, t}}+\gamma\rho(Z^{c}(o_{\agentNum,t+1}))-\rho(Z^{c}(o_{\agentNum,t}))$. To effectively optimize (\ref{obj: RS-CRL}) by updating the Lagrange multipliers, we design a multi-agent constrained policy gradient algorithm 
to update the policy $\pi_{\agentNum,t}(a_{i,t}|\observation_{\agentNum,t})$ under the CTDE framework\cite{yu2022mappo} (see Algorithm \ref{alg:CCE-MAPPO}). Implementation details of the parameter setting can be found in the Supplementary Material. We construct actor models with parameters $\theta$ and critic models with parameters $\phi$ using MLPs. The agent $i$'s actor network is trained to maximize:
\begin{equation}
\begin{aligned}
    &L^{\text{CLIP}}(\theta_{\agentNum})=\min\Big[(\frac{\pi_{\theta_{\agentNum}}(\cdot)}{\pi_{\theta^{old}_{\agentNum}}(\cdot)}(A_{\agentNum}^r-\lambda(A_{\agentNum}^c-\beta)),\\
    &\operatorname{clip}(\frac{\pi_{\theta_{\agentNum}}(\cdot)}{\pi_{\theta^{old}_{\agentNum}}(\cdot)},1-\eta_3,1+\eta_3)(A_{\agentNum}^r-\lambda(A_{\agentNum}^c-\beta))\Big]\label{obj:cmappo-clip}
\end{aligned}
\end{equation}

The critic network is trained to minimize the loss function:
\begin{align}
L^{VF}=\|V^{\pi_\agentNum,\boldsymbol{\pi}_{-i}}_{\phi_{\agentNum}}(s)-R_\agentNum\|_{2}^{2}
\label{loss: VF}
\end{align}
where $R$ is the discounted reward-to-go.

For the risk sensitivity section, we apply SplineDQN\cite{LuoLDSP22splinedqn} to estimate the cost value of the observations. Using risk sensitivity optimization with the CVaR method, we can adjust the confidence level $\alpha$ to control whether the policy exhibits risk-seeking or risk-avoidance behavior. In our algorithm, a high value of $\alpha$ reflects a conservative risk preference, where the agent is particularly focused on mitigating the extreme tail of the cost distribution (e.g., severe collision risks). Conversely, a low value of $\alpha$ indicates a risk-seeking behavior, with the agent prioritizing average returns while neglecting extreme tail risks. A higher $\alpha$ leads to more conservative driving where the $\beta$ constraint is largely satisfied, while a lower $\alpha$ can result in more aggressive behaviors that risk violating the safety constraint $\beta$.

To satisfy the requirements of Definition~\ref{def:Offline-MA-RL}, we pre-train a world model $\mathcal{M}$ using real-world driving datasets to represent various autonomous driving scenarios, as shown in Equation \ref{eqn:imitate-policy}. The world model serves as the environment for solving the CCE in the subsequent MARL process. We utilize objective \ref{obj:marl-kl-mmd} to constrain the agents' policies during the CCE capture, ensuring that they reference human driving strategies, which helps generate scenarios with a degree of realism. Then, we apply objective \ref{obj: RS-CRL} to capture CCEs of varying intensities, thereby achieving the goal of generating safety-critical scenarios. During the generation of these critical scenarios, the regularization term from objective \ref{obj:marl-kl-mmd} ensures that the generated scenarios closely approximate real-world driving situations.

\begin{algorithm}[!t]
   \caption{TrafficGamer for capturing CCE}
   \label{alg:CCE-MAPPO}
\begin{algorithmic}
    \STATE {\bfseries Input:} 
    Offline dataset $\dataset_l$, the number of total agents $\MakeUppercase{\agentNum}$, constraint threshold $\beta$, Lagrange multiplier $\lambda$, rollout rounds $B$,
    update rounds $J$, distributional cost value critic $\{Z^c_\agentNum\}_{\agentNum=1}^{\MakeUppercase{\agentNum}}$, the policies $\{\pi_{\theta_{\agentNum}}\}_{\agentNum=1}^{\MakeUppercase{\agentNum}}$, the reward value critic $\{V_{\phi_i}\}_{\agentNum=1}^{\MakeUppercase{\agentNum}}$
    \STATE {\bfseries Output:} 
    $\{\pi^{CCE}_{\theta_{\agentNum}}\}_{\agentNum=1}^{\MakeUppercase{\agentNum}}$
    \STATE Initialize the world model $\mdp^l$,  observation $\{o_{\agentNum,0}\}_{\agentNum=1}^{\MakeUppercase{\agentNum}}$ from POCMDP and the roll-out dataset $D_{roll}$;
    \STATE Pre-train the world model $\mdp_\theta^l$ with (Section~\ref{subsec:world-model});

    \FOR{$b=1,2,\dots,B$}
    \FOR{$i=1,2,\dots,I$}
    \STATE Perform roll-out with the policy $\boldsymbol{\pi}_\theta $ in a scenario;
    \STATE Collect trajectories for each agent $\tau_{\agentNum,b} = [o_{\agentNum,0} ,a_{\agentNum,0}, r_{\agentNum,0}, c_{\agentNum,0},..,o_{\agentNum,T}, a_{\agentNum,T}, r_{\agentNum,T}, c_{\agentNum,T}]$;
    \STATE Calculate reward advantages $A^{\boldsymbol{\pi},r}_{\agentNum,t}$, cost advantages $A_{\agentNum,t}^{\boldsymbol{\pi},c}$ and total rewards $R_{\agentNum, t}$ with (\ref{obj:marl-pg}) from the trajectory and add these terms to the dataset $D_{roll}$;
    \ENDFOR
    \ENDFOR
    \FOR{$j=1,2,\dots,J$}
    \FOR{$\agentNum=1,2,\dots,\MakeUppercase{\agentNum}$}
    
     \STATE Sample data points from $D_{roll}$ for each agent;
    \STATE update the value function $L^{VF}$ with (\ref{loss: VF});
    \STATE Update policy function by minimizing the loss: $-L^{CLIP}-\eta\mathcal{H}(\pi_{\agentNum,j})$ ($-L^{CLIP}$ refers to (\ref{obj:cmappo-clip}));
    \STATE Update the cost distribution $Z^c_\agentNum$ by distributional Bellman operator with the equation (\ref{obj: dbe});
    \ENDFOR
    \STATE Update the Lagrange multiplier by minimizing the loss: ${L^{\lambda}\colon\lambda[\mathbb{E}_{\mathcal{D}_{roll}}(c_i)-\beta]}$;
    
    \ENDFOR

\end{algorithmic}
\end{algorithm}


\section{Experimental Results}
\label{sec: experimental results}
{\bf Dataset.} We evaluate our model using the publicly available Argoverse 2\cite{wilson2021argoverse2} dataset, which includes 250,000 scenarios from six urban environments in the U.S. The first 5 seconds are the observed window, while the next 6 seconds are the forecasted horizon. To validate our approach, we also assess with the Waymo Open Motion Dataset (WOMD)\cite{Ettinger2021waymodataset}. This dataset features over 100,000
scenes, each lasting 20 seconds and recorded at 10 Hz. The first 3 seconds are the observed window, while the next 8 seconds are the forecasted horizon.

{\bf Experimental Setting.} Based on the trajectory prediction task setting, our generative world model predicts an agent's future states by observing the agent's historical information, map features, surroundings, and neighboring agents. The performance of the pre-training world model can be checked in the Supplementary Material. By iteratively performing this prediction process, TrafficGamer teaches a CCE solver to generate traffic congestion scenarios with varying degrees of competition. To verify the generative performance of our algorithm, we divide the scenarios into six types, including (1) {\it Merge} where two separate lanes of traffic join into a single lane, (2) {\it Dual-lane intersection} where five cars with different destinations drive through a two-way intersection, (3) {\it T-junction} where one road ends at a perpendicular junction with another road, forming a "T" shape, (4) {\it Dense-lane intersection} where cars enter a four-way intersection with dense traffic. (5) {\it Roundabout} where traffic flows counterclockwise around a central circle, and (6) {\it Y-junction} where three directions of traffic flow converge at a single intersection.




{\bf Metrics.} We leverage a comprehensive set of statistical metrics to evaluate the ﬁdelity and effectiveness of the proposed \md. The following metrics are included:

\textit{Fidelity} metrics measure how well the simulated traffic distribution (i.e., the distribution of vehicles' temporal and spatial features) matches the observed data distribution.
We implement these fidelity metrics with \textit{Hellinger distance} $D_H$, \textit{Kullback-Leibler divergence} (KL) $D_{KL}$, \textit{Wasserstein distance} $D_{W}$ between simulated probability distribution and real-world probability distribution. We demonstrate the differences in velocity, acceleration, vehicle distance, and steering angle.

\textit{Exploitability} metric reflects the optimality of the joint policies under a GS-MG. The traffic simulation reflects the optimality of simulated vehicles in quickly reaching their destinations under traffic rules and other constraints.
Under Definition \ref{def:cce-gap}, a $\text{CCE-gap}(\agentNum) =V^{\dagger,\boldsymbol{\pi}_{-\agentNum}}_{\agentNum,0}(s)-V^{\boldsymbol{\pi}}_{\agentNum,0}(s)$ quantifies the deviation between the learned policies of each agent and the performance of the best equilibrium policy. We assess the CCE-gap, as it better aligns with the decision-making processes of human drivers, who typically base their actions on the behaviors of nearby vehicles.

\textit{Diversity} metric refers\cite{zhang2025DriveGen} to measuring the diversity of trajectories generated by the vehicles controlled by each method. A trajectory consists of the sequence of position coordinates along a vehicle's path. The generated trajectories from each scenario are randomly split into pairs, and the average pairwise differences within each pair are calculated, which serves as the diversity metric. We only consider algorithm-controlled vehicles.

In the following experimental tables, the values highlighted in \textcolor{blue}{blue} represent the best performance, while those in \textcolor{green}{green} represent the second-best performance.

{\bf Baselines.} We compare the proposed method with supervised learning method QCNet\cite{zhou2023qcnet}, Donut\cite{knoche2025donut}, RL method Multi-Agent Proximal Policy Optimization (MAPPO)\cite{yu2022mappo, Kuba2022HAPPO}, self-play RL\cite{Francis2025selfplay, kazemkhani2025gpudrive}, RS-Pluto\cite{peng2024rs}, and game-theoretical method GameFormer\cite{Huang2023gameformer}. QCNet jointly predicts the trajectory of multiple agents under a supervised learning framework. Donut encodes historical trajectories and predicts future trajectories with an autoregressive model. MAPPO is a multi-agent policy gradient algorithm designed for MARL; in this implementation, we adopt the HAPPO paradigm with heterogeneous agents. RS-Pluto adopts a hybrid IL+RL interactive trajectory prediction paradigm. Self-play RL enables the agent to generate abundant training data, thereby enabling continuous learning and strategy refinement. GameFormer proposes a game-theoretic model and learning framework for interactive prediction and planning. The QCNet and MAPPO methods also serve as ablation baselines.

\begin{figure*}[htbp]
\centering
\centerline{\includegraphics[width=2\columnwidth]{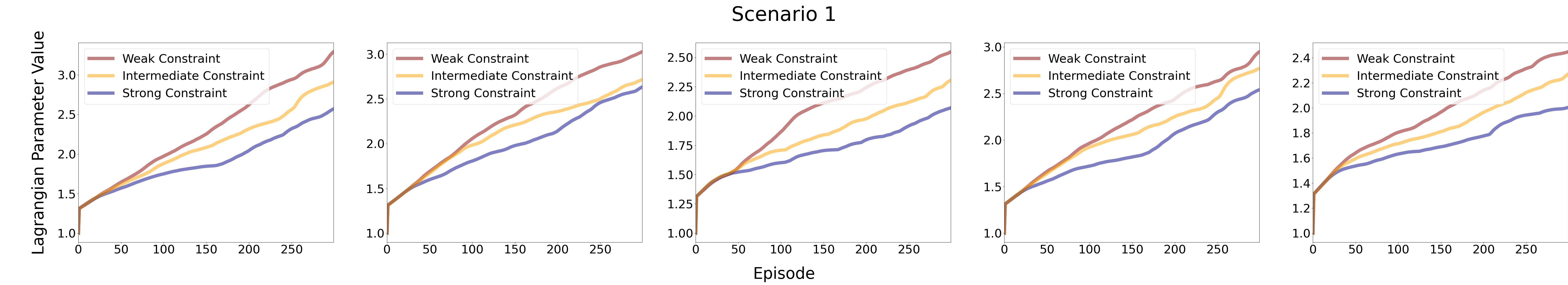}}
\caption{{\bf Variation in Lagrangian parameters under varying constraints}. Each column corresponds to one of the agents in the multi-agent environment.}
\label{fig: lagrangian_parameter}
\end{figure*}

The supplementary material contains a description of the training details.


\begin{figure*}[htbp]

    \begin{minipage}[t]{0.33\textwidth}
        \centerline{\includegraphics[width=\columnwidth]{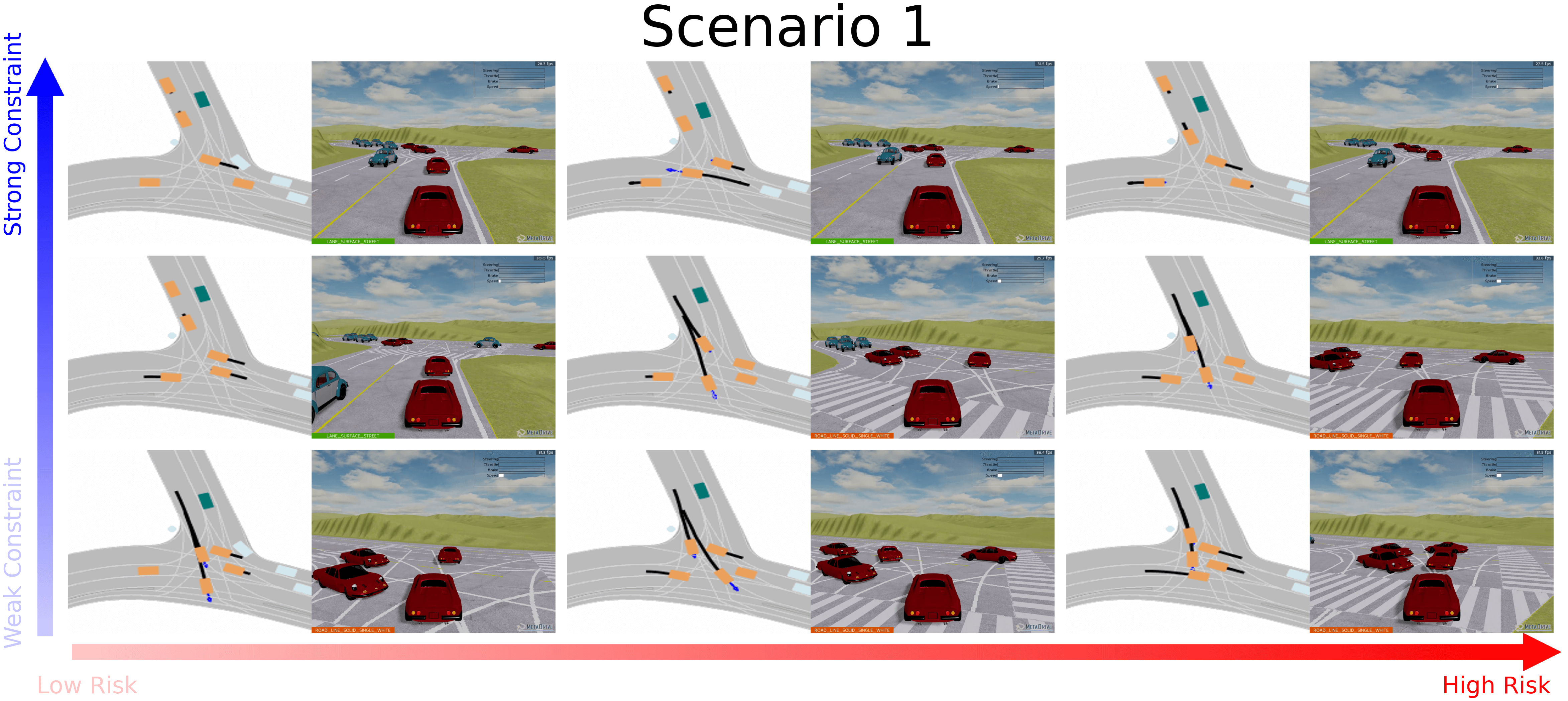}}
    \end{minipage}
    \begin{minipage}[t]{0.33\textwidth}
        \centerline{\includegraphics[width=\columnwidth]{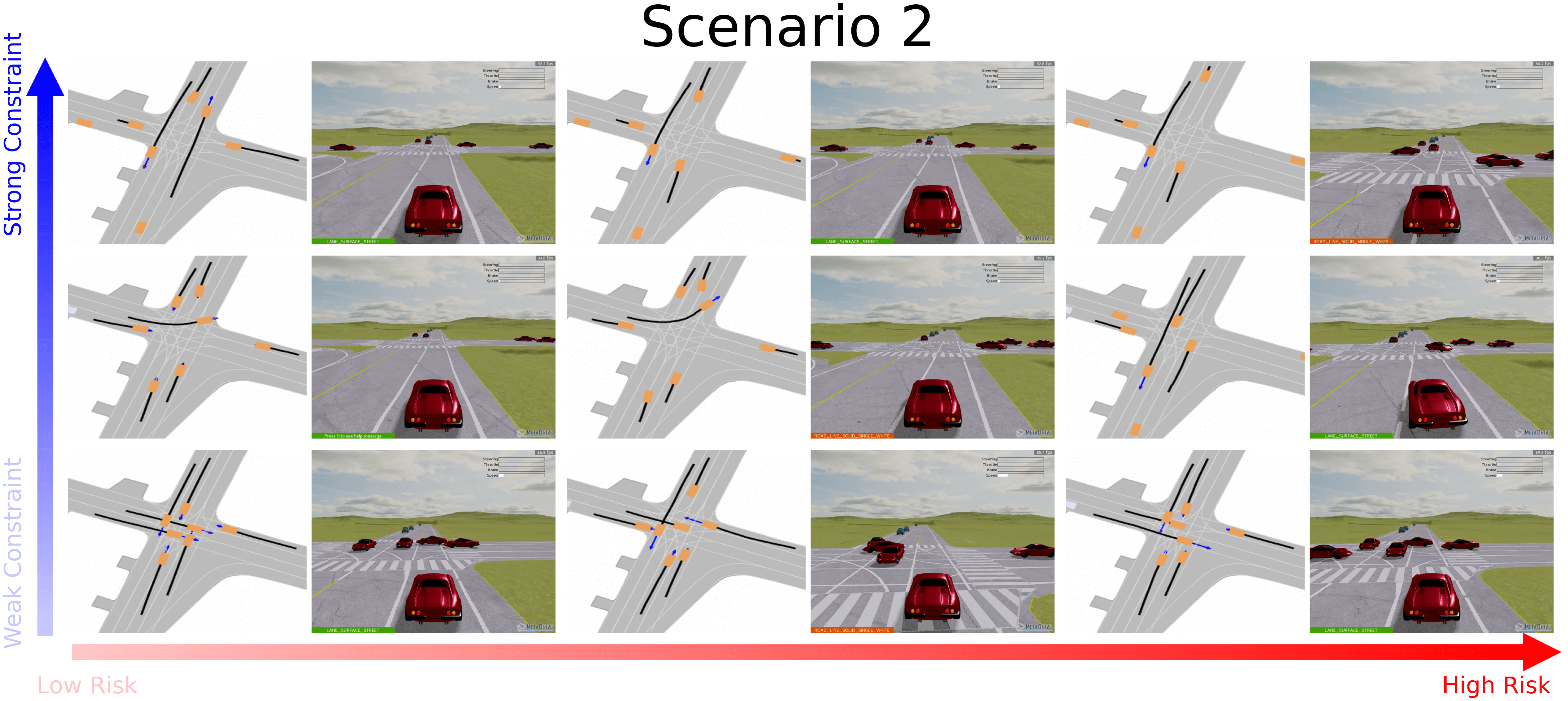}}
    \end{minipage}
    \begin{minipage}[t]{0.33\textwidth}
        \centerline{\includegraphics[width=\columnwidth]{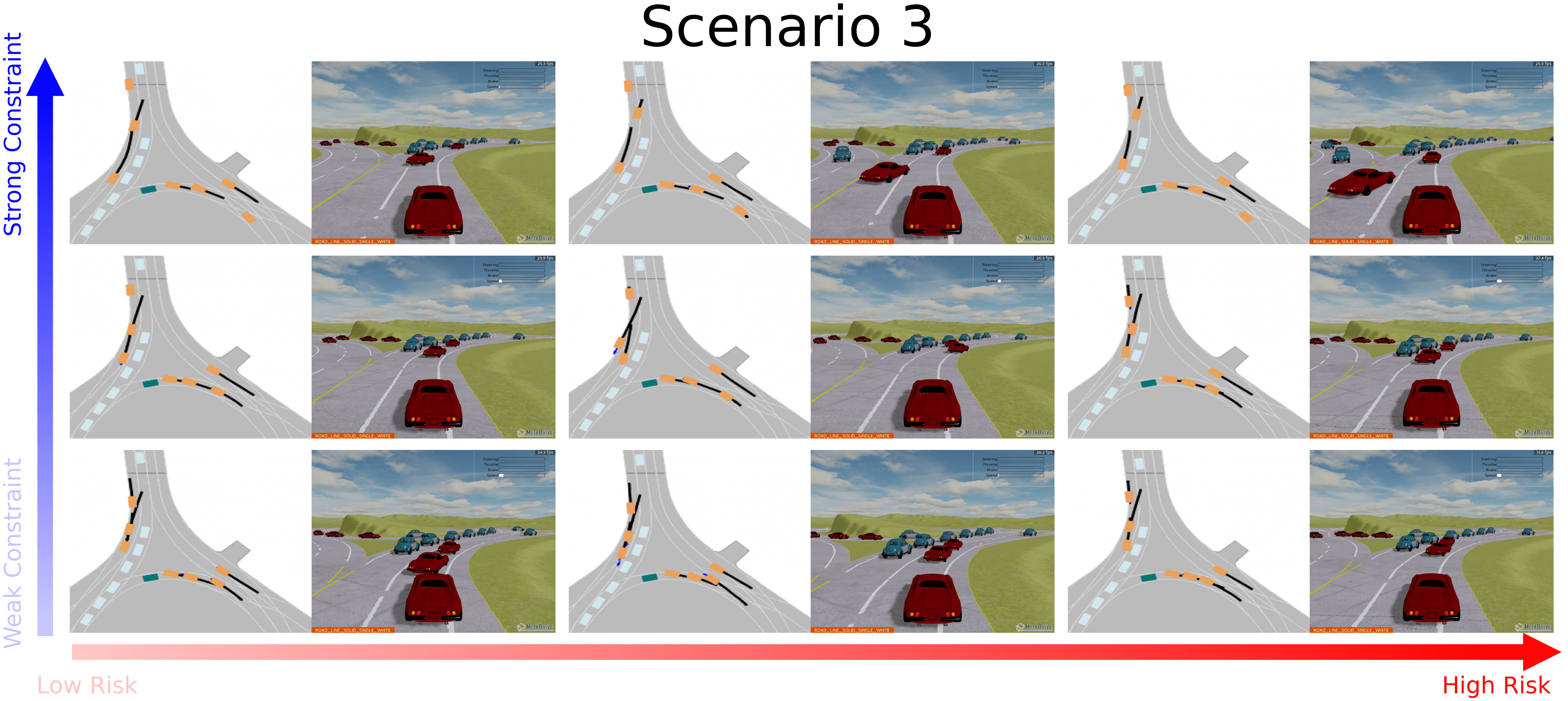}}
    \end{minipage}

    \begin{minipage}[t]{0.33\textwidth}
       \centerline{\includegraphics[width=\columnwidth]{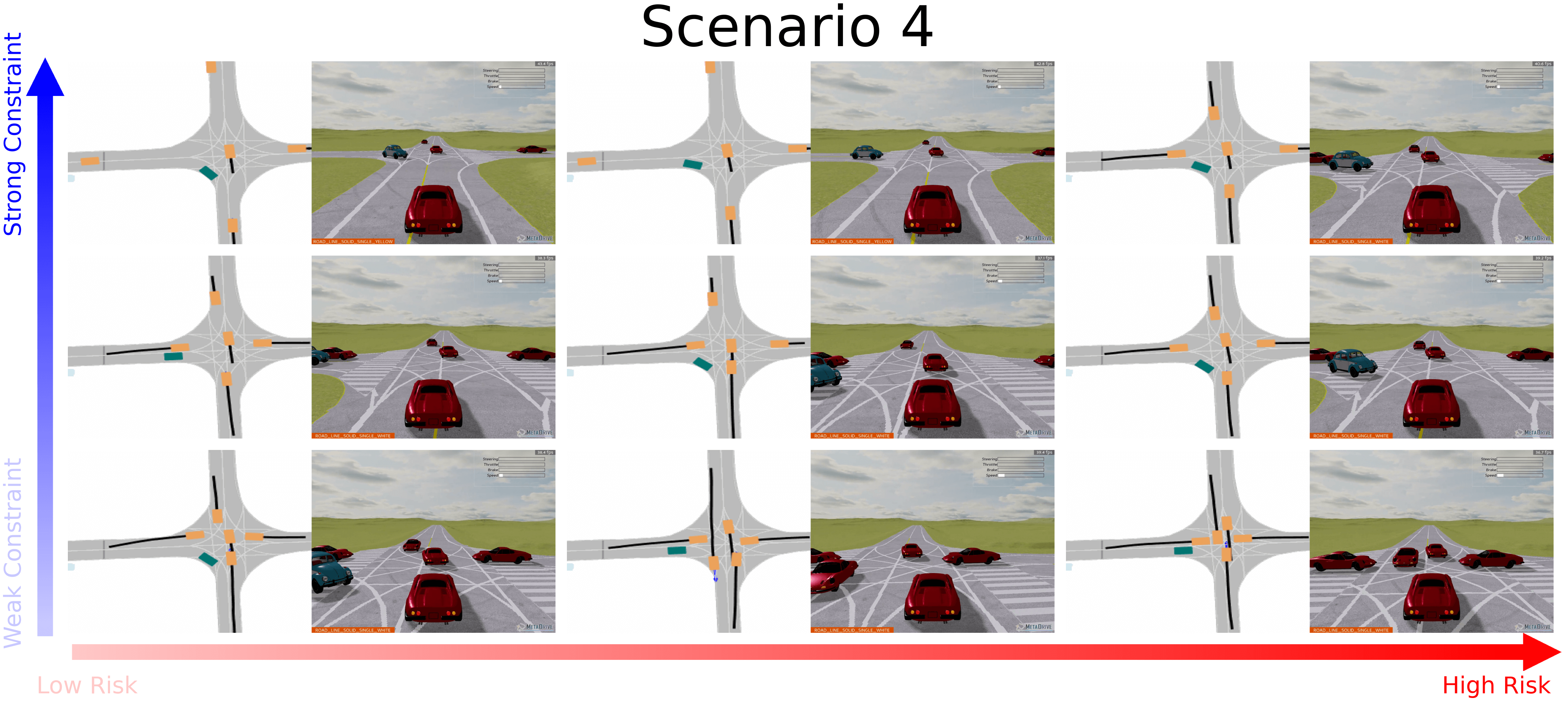}}
    \end{minipage}
    \begin{minipage}[t]{0.33\textwidth}
    \centering
        \centerline{\includegraphics[width=\columnwidth]{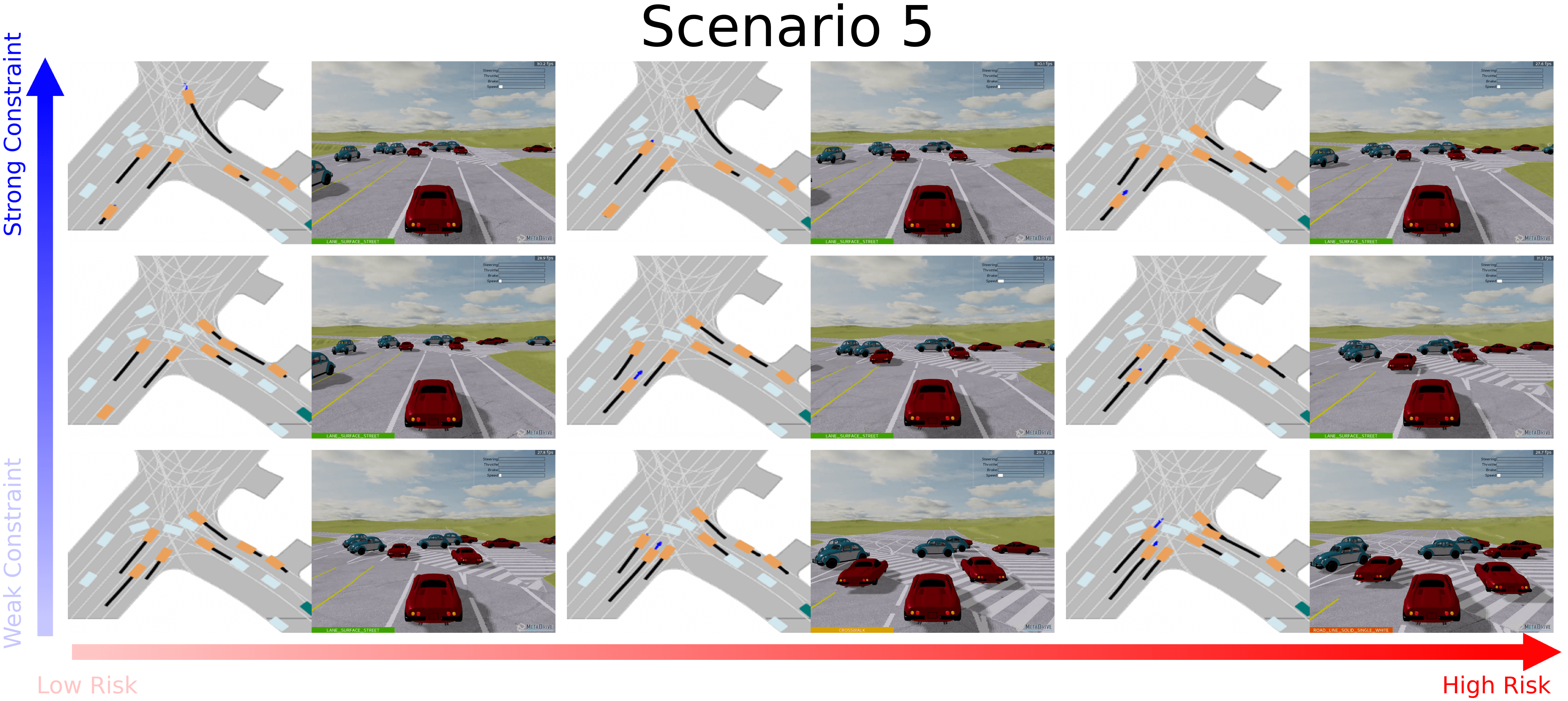}}
    \end{minipage}
    \begin{minipage}[t]{0.33\textwidth}
     \centerline{\includegraphics[width=\columnwidth]{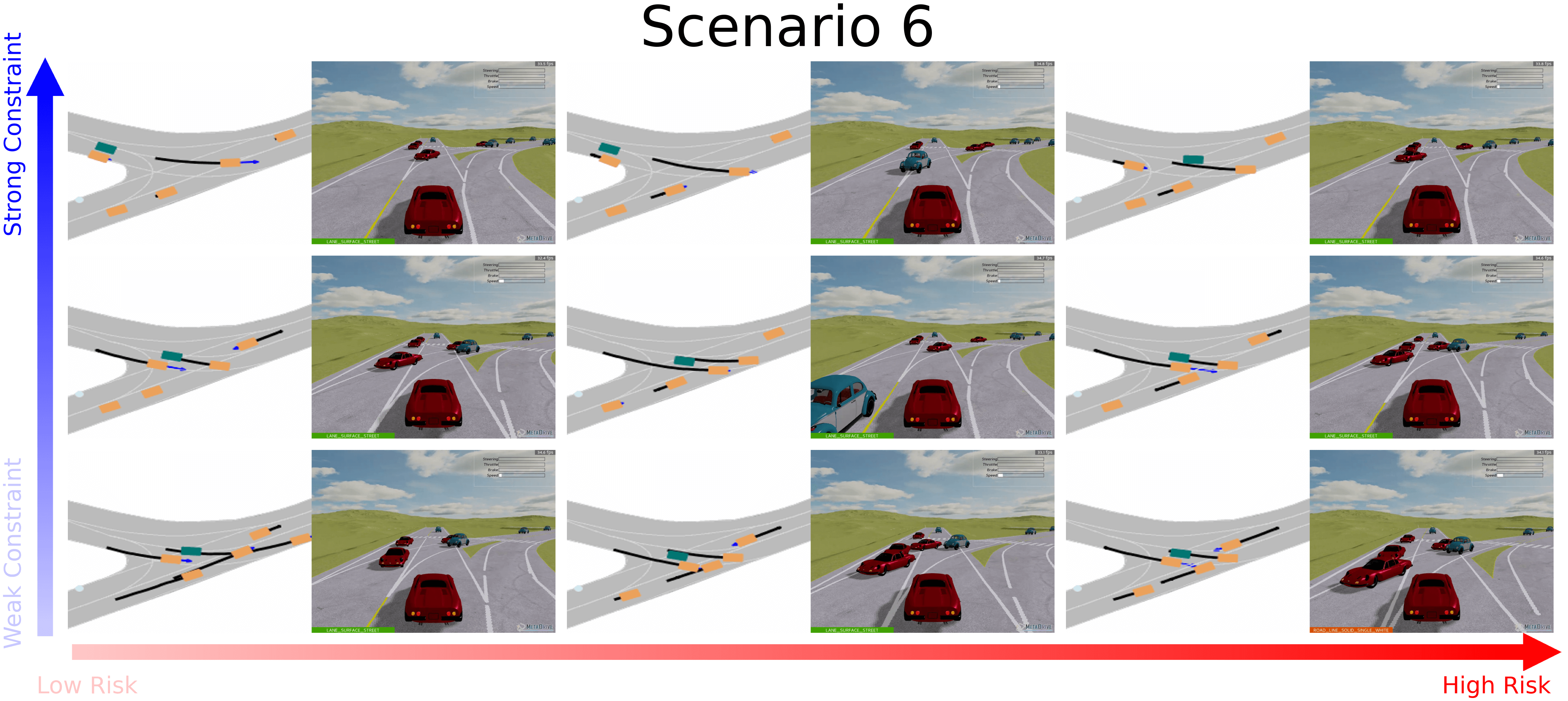}}
    \end{minipage}
    \caption{{\bf Visualization of flexible generated trajectories (Argoverse 2).} Visualization of all Argoverse 2 scenarios in a 3x3 grid layout with 2D and 3D simulations. In each grid, the distance constraint relaxes from top to bottom, and the risk level strengthens from left to right. In the 2D scenes, orange cars are algorithm-controlled, green cars and gray cars are environmental vehicles, black lines are vehicle trajectories, and blue arrows indicate travel direction. In the 3D scenes, red cars are algorithm-controlled, while blue cars represent environmental vehicles.}
    \label{fig: simulation1}
\end{figure*}

\subsection{Simulation of Diverse Safety-Critical Scenarios}
To demonstrate the ability of \md\;to generate diverse traffic scenarios, we select some examples of different road types and visualize the scenarios generated with varying degrees of competitive behavior.
Automated driving scenarios involve complex elements including road structures, traffic regulations, and vehicle behaviors, which can be challenging to interpret directly from raw data. To address this, we simulate the actual behavior of vehicles in various safety-critical scenarios using 2D and 3D visualizations. These are captured in third-person and first-person views, to provide a clearer understanding of the dynamics. For 2D visualization, we simply use specific colors and shapes based on the Python Matplotlib library to represent traffic scenes. To capture the real-world complexity of traffic behaviors more accurately, \md\;enables large-scale 3D traffic scenario modeling and simulation through the MetaDrive simulator\cite{Li2023metadrive}. 

In this experiment, we explore how well \md\;can generate different safety-critical scenarios under 6 random distinct traffic scenarios. To generate diverse traffic scenarios, we design a constrained and risk-sensitive policy optimization method that captures the equilibrium under different levels of tightness. Specifically, based on the objective \ref{obj: RS-CRL}, we adjust the inter-vehicle distance constraint $\beta$, and the risk coefficient $\alpha$, to derive a diverse set of traffic scenarios under various game contexts. The $\beta$ defines the minimum physical distance that must be maintained between vehicles and serves as a binding constraint in the optimization problem. Meanwhile, $\alpha$ characterizes the agent's risk sensitivity. 




By comparing the scenarios generated with different levels of inter-vehicle distance constraints in Fig. \ref{fig: simulation1} (from bottom to top), we find that, as the inter-vehicle distance $\beta$ increases, our model adapts, leading to traffic scenarios characterized by less competitive behavior and safer navigation in all examined situations. This outcome arises because maintaining distance constraints requires the cooperation of multiple vehicles. Imposing more restrictive constraints (i.e., increasing the distance) significantly enhances the impact of this cooperation on the optimization of Objective (\ref{obj: CRL}) (dynamically controlled by the Lagrange parameter $\lambda$). 
Fig. \ref{fig: lagrangian_parameter} displays the variation in the Lagrangian penalty factor during the training process. As the constraints become tighter, the Lagrangian penalty term also increases, indicating that it has a larger impact on Objective (\ref{obj: CRL}). On the other hand, when we reduce the required distance, agents begin to prioritize their interests, which significantly increases the likelihood of traffic congestion where no agent can further optimize their policy.

Similarly, by comparing the scenarios generated with different $\alpha$ levels of risk sensitivity in Fig. \ref{fig: simulation1} (from left to right), we find that imposing a lower confidence level $\alpha$ leads to more aggressive and high-risk driving behaviors. A higher confidence level forces the agent to satisfy the constraints with greater probability, resulting in driving strategies that feature lower speeds, increased spacing between vehicles, and more careful navigation through complex traffic scenarios. However, setting a lower confidence level results in more aggressive and high-risk driving behaviors, characterized by faster vehicle speeds and shorter follow-up distances. The system's tolerance for some aggressive behaviors (such as overtaking, etc.) also increases. These results demonstrate how variations in risk sensitivity and distance constraints impact vehicle behavior in real-world industrial driving scenarios.

\begin{table}[htbp]
\centering
\caption{Diversity evaluation (higher is better)}\label{table:diversity-results}
\resizebox{0.5\textwidth}{!}{
\begin{tabular}{c|ccccccc}
\hline
\multicolumn{1}{c|}{\textbf{Method}} & \textbf{Scenario 1} & \textbf{Scenario 2} & \textbf{Scenario 3} & \textbf{Scenario 4} & \textbf{Scenario 5} & \textbf{Scenario 6} & \textbf{Avg} \\ \hline
QCNet      & 0.31 & 0.81 & 0.05 & 0.37 & 0.13 & 0.86 & 0.42 \\
Donut      & 0.15 & 0.49 & 0.10 & 0.31 & 0.19 & \cellcolor{green!25}1.68 & 0.49 \\
GameFormer & 0.33 & 1.11 & \cellcolor{blue!25}0.67 & 0.29 & 0.27 & 0.47 & 0.52 \\
MAPPO      & 0.11 & \cellcolor{green!25}1.50 & 0.16 & 0.16 & 0.09 & 0.15 & 0.36 \\
Self-Play RL & 0.18 & 0.26 & 0.27 & 0.11 & \cellcolor{blue!25}0.65 & \cellcolor{blue!25}1.98 & \cellcolor{green!25}0.58 \\
RS-Pluto   & \cellcolor{green!25}0.34 & 0.18 & 0.21 & \cellcolor{green!25}1.31 & 0.03 & 0.67 & 0.46 \\
TrafficGamer & \cellcolor{blue!25}1.05 & \cellcolor{blue!25}5.79 & \cellcolor{green!25}0.45 & \cellcolor{blue!25}3.27 & \cellcolor{green!25}0.41 & 1.05 & \cellcolor{blue!25}2.0 \\ \hline
\end{tabular}
}
\end{table}

To fully validate the effects of risk sensitivity and distance constraints, we employ a diversity metric to evaluate the diversity of trajectories generated by various methods. Based on the diversity of evaluation results presented in Table \ref{table:diversity-results}, TrafficGamer demonstrates a significant advantage in generating diverse trajectories compared to other methods. It achieves the highest diversity score in four out of the six scenarios (Scenario 1, 2, and 4) and the highest overall average score of 2.0, substantially higher than the second-best average of 0.58. These results indicate that TrafficGamer excels at generating a wide variety of trajectories, demonstrating its superior ability to maintain diversity across different traffic scenarios. Furthermore, to effectively demonstrate the varying risk levels of the generated scenarios, we estimate vehicle collision times within specified spatial-temporal ranges.

\begin{figure*}[htbp]
\centering
\centerline{\includegraphics[width=2\columnwidth]{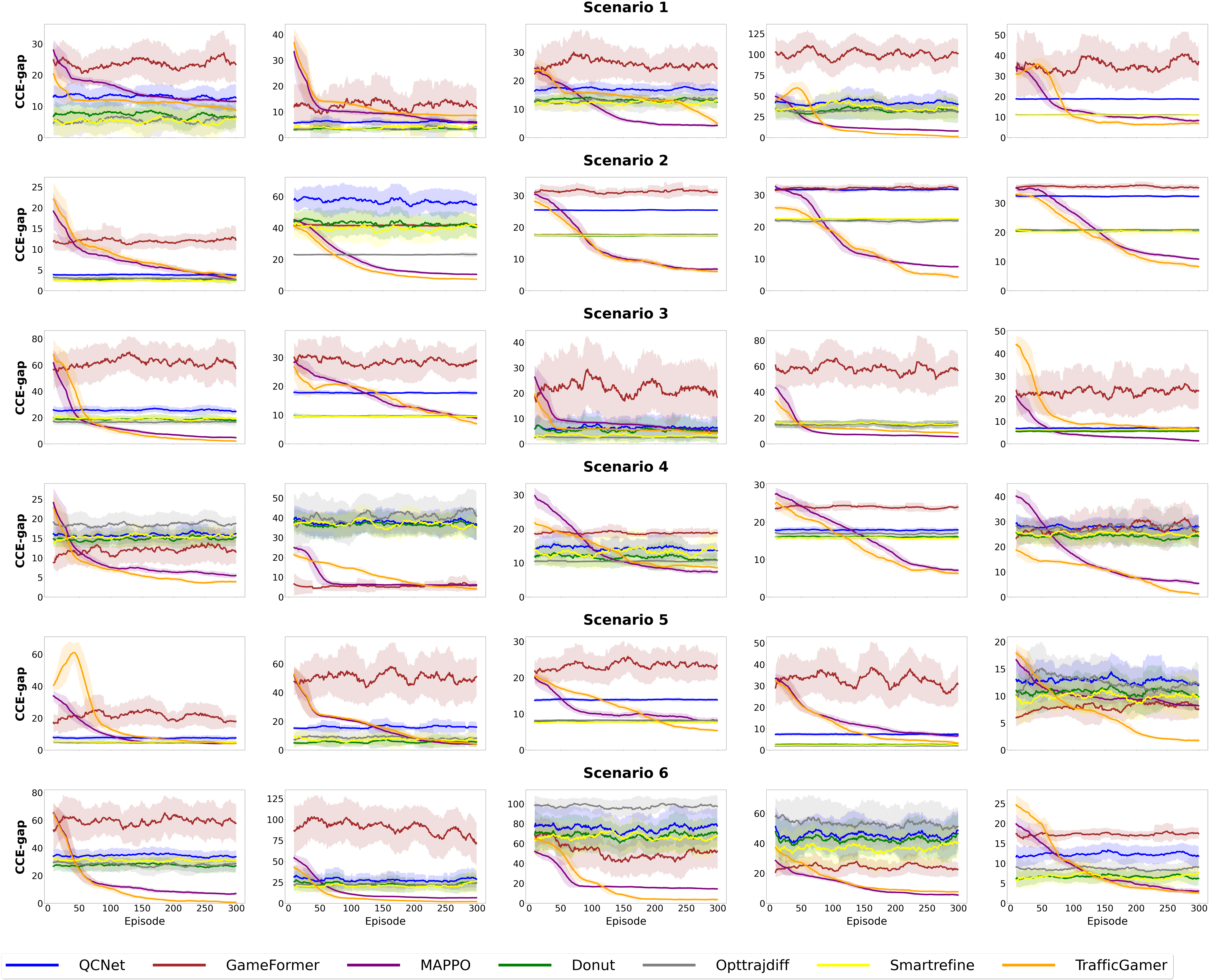}}
\caption{{\bf CCE-gap obtained from the Breaking the Equilibrium approach}. Each row
represents a scenario and scenarios 1-6 correspond to {\it Y-Junction},
{\it Dense-lane intersection},
{\it Roundabout},
{\it Dual-lane intersection},
{\it T-Junction}, and
{\it Merge} respectively. Each column
corresponds to one of the agents in the multi-agent environment. Because of the space limitations, we only showcase the 5 vehicles' CCE-gap of each scenario.}
\label{fig: cce-gap1}
\end{figure*}

\subsection{Efficient Exploitability Under a Variety of Scenarios}

Standard RL algorithms focus on reward maximization; however, in the multi-agent autonomous driving environment, we primarily consider exploitability\cite{Sokota2023MMD}, which measures the extent to which a vehicle's policy can exploit the current traffic policies. An ideal driving equilibrium should have zero exploitability, meaning no single vehicle can achieve greater benefits by continually improving the policy. In this paper, we follow the methodology outlined in\cite{li2024CMD} and utilize the \textbf{CCE-gap} (see Definition~\ref{def:cce-gap}) to measure exploitability. 
Unlike the commonly studied toy environments (e.g., Bargaining, TradeComm, and Battleship\cite{li2024CMD}), the traffic scenarios involve complex game contexts, multiple agents, and continuous action space, which makes the best response $\pi^{\dagger}_\agentNum$ computationally intractable, thus
accurately calculating the exact CCE-gap becomes challenging.
Therefore, we estimate the CCE-gap with different seeds by empirically approximating $\pi^{\dagger}_\agentNum$ via "Breaking the Equilibrium". Specifically, upon the convergence of the studied algorithm, we estimate each agent's best response under the current equilibrium, denoted as $\pi^{\dagger}_\agentNum$, by fixing the policies of the other $I-1$ agents and continuously encouraging this agent to maximize its current reward. This process assesses the agent's ability to disrupt the existing equilibrium. If none of the policies can achieve significantly higher rewards, it indicates that the experimental algorithm has successfully identified a reliable CCE.

We compare the performance of TrafficGamer in all Argoverse 2 scenarios with other baselines. Fig. \ref{fig: cce-gap1} and Table \ref{table:cce-gap-results} illustrate the CCE-gaps for partial agents during training. The training curves of supervised learning methods indicate unstable performance and difficulty converging at the CCE. The results illustrate that end-to-end imitation learning methods cannot model competitive interactions among agents, making it difficult to optimize policies that capture CCE. MAPPO outperforms other baselines but falls short of the results achieved by TrafficGamer.
MAPPO faces challenges in efficiently exploring the entire policy space of the multi-agent game environment during optimization, making it difficult to capture the underlying CCE. When faced with different road types, Self-Play RL and RS-Pluto perform adequately in some scenarios but lack stability. For example, in Scenario 3 of Fig. \ref{fig: cce-gap1}, neither algorithm converges well, and some agents even deviate further from the equilibrium. As a CCE solver, TrafficGamer refers to CCE-V-Learning, and MMD, ensuring that agents' policies are distributionally aligned with human-driven policies, support stable exploration, and cover all kinds of road map $\mathcal{M}$. This allows each agent to converge to an approximate CCE gradually.

\begin{table}[htbp]
\centering
\caption{CCE-gaps across six scenarios (lower is better).}
\label{table:cce-gap-results}
\resizebox{0.5\textwidth}{!}{
\begin{tabular}{l|cccccc}
\hline
\textbf{Method} & \textbf{Agent1} & \textbf{Agent2} & \textbf{Agent3} & \textbf{Agent4} & \textbf{Agent5} & \textbf{Avg} \\
\hline
\multicolumn{7}{c}{\textbf{Scenario 1}} \\ \hline
QCNet                  & 10.72${\pm}$4.19 & 3.99${\pm}$0.99  & 16.58${\pm}$1.95 & 40.00${\pm}$9.26  & 18.65${\pm}$0.45  & 17.99${\pm}$2.08 \\
GameFormer             & 21.38${\pm}$5.87 & 9.41${\pm}$7.07 & 24.33${\pm}$5.75 & 101.20${\pm}$17.26 & 36.93${\pm}$10.03 & 38.65${\pm}$4.54 \\
MAPPO                  & 9.59${\pm}$0.47 & 3.68${\pm}$0.71  & \cellcolor{blue!25}4.24${\pm}$0.42 & \cellcolor{green!25}7.87${\pm}$0.76  & 8.29${\pm}$0.80   & \cellcolor{green!25}6.73${\pm}$0.29  \\
Donut                  & \cellcolor{green!25}4.37${\pm}$3.31  & \cellcolor{blue!25}1.44${\pm}$1.19  & 12.46${\pm}$2.30 & 31.14${\pm}$13.47 & 10.86${\pm}$0.34  & 12.05${\pm}$2.82 \\
Self-Play RL           & \cellcolor{blue!25}3.05${\pm}$1.55 & \cellcolor{green!25}4.23${\pm}$2.85& 8.40${\pm}$0.83& 19.96${\pm}$4.57& \cellcolor{green!25}8.07${\pm}$1.78  & 8.74${\pm}$1.19 \\
RS-Pluto               & 11.12${\pm}$1.67 & 18.28${\pm}$4.17& 17.03${\pm}$3.18& 52.71${\pm}$19.82& 15.61${\pm}$1.74  & 22.95${\pm}$4.13 \\
TrafficGamer           & 6.73${\pm}$0.60 & 6.53${\pm}$0.23 & \cellcolor{green!25}4.97${\pm}$1.27 & \cellcolor{blue!25}1.41${\pm}$0.41 & \cellcolor{blue!25}6.85${\pm}$0.86 & \cellcolor{blue!25}5.30${\pm}$0.34 \\
\hline
\multicolumn{7}{c}{\textbf{Scenario 2}} \\ \hline
QCNet                  & 3.78${\pm}$0.50  & 54.74${\pm}$7.13 & 25.39${\pm}$0.16 & 31.69${\pm}$0.31 & 32.32${\pm}$0.32 & 29.58${\pm}$1.43 \\
GameFormer             & 12.23${\pm}$2.43 & 41.95${\pm}$0.74 & 31.04${\pm}$1.21 & 32.13${\pm}$0.89 & 35.28${\pm}$0.93 & 30.53${\pm}$0.62 \\
MAPPO                  & 2.64${\pm}$0.58 & 10.38${\pm}$0.49 & 6.79${\pm}$0.45  & 7.51${\pm}$0.32  & 10.82${\pm}$0.33 & 7.63${\pm}$0.20  \\
Donut                  & \cellcolor{green!25}2.58${\pm}$0.53  & 40.31${\pm}$7.34 & 17.32${\pm}$0.14 & 22.42${\pm}$0.24 & 20.34${\pm}$0.25 & 20.59${\pm}$1.47 \\
Self-Play RL           & \cellcolor{blue!25}1.88${\pm}$1.77 & \cellcolor{blue!25}5.83${\pm}$0.99& \cellcolor{blue!25}4.45${\pm}$1.18& \cellcolor{blue!25}3.67${\pm}$1.04& \cellcolor{blue!25}1.87${\pm}$1.05  & \cellcolor{blue!25}3.54${\pm}$0.55 \\
RS-Pluto               & 32.28${\pm}$5.12 & 36.72${\pm}$4.98& 29.61${\pm}$3.83& 32.53${\pm}$5.84& 33.78${\pm}$4.73  & 32.98${\pm}$2.21 \\
TrafficGamer           & 3.41${\pm}$0.25 & \cellcolor{green!25}7.28${\pm}$0.49 & \cellcolor{green!25}6.05${\pm}$0.52 & \cellcolor{green!25}4.31${\pm}$0.52 & \cellcolor{green!25}8.22${\pm}$0.68 & \cellcolor{green!25}5.85${\pm}$0.23 \\
\hline
\multicolumn{7}{c}{\textbf{Scenario 3}} \\ \hline
QCNet                  & 24.61${\pm}$2.27 & 17.63${\pm}$0.69 &4.24${\pm}$4.51  & 15.66${\pm}$2.52 & 6.98${\pm}$0.72  & 13.82${\pm}$1.15  \\
GameFormer             & 57.38${\pm}$14.66& 29.00${\pm}$6.30 & 16.35${\pm}$10.87& 56.73${\pm}$12.84& 23.48${\pm}$7.78 & 36.59${\pm}$4.89 \\
MAPPO                  & \cellcolor{green!25}4.68${\pm}$0.58  & 8.92${\pm}$0.80  & 3.37${\pm}$0.55  & \cellcolor{blue!25}5.47${\pm}$0.50  & \cellcolor{blue!25}1.33${\pm}$0.47 & \cellcolor{blue!25}4.75${\pm}$0.26  \\
Donut                  & 17.78${\pm}$2.08 & 9.43${\pm}$0.23  & \cellcolor{green!25}2.81${\pm}$4.54  & 15.34${\pm}$2.52 & \cellcolor{green!25}5.69${\pm}$0.60  & 10.21${\pm}$1.13 \\
Self-Play RL           & 21.05${\pm}$4.98 & \cellcolor{green!25}7.32${\pm}$1.32& 23.76${\pm}$12.10& 24.87${\pm}$6.35& 15.35${\pm}$2.36  & 18.47${\pm}$2.96 \\
RS-Pluto               & 21.65${\pm}$11.56 & 19.47${\pm}$7.14& 19.07${\pm}$7.54& 41.85${\pm}$11.82& 18.56${\pm}$7.85  & 24.12${\pm}$4.21 \\
TrafficGamer           & \cellcolor{blue!25}2.06${\pm}$0.46 & \cellcolor{blue!25}7.00${\pm}$0.74 & \cellcolor{blue!25}2.24${\pm}$0.44 & \cellcolor{green!25}8.25${\pm}$0.44 & 6.51${\pm}$0.64  & \cellcolor{green!25}5.21${\pm}$0.25 \\
\hline
\multicolumn{7}{c}{\textbf{Scenario 4}} \\ \hline
QCNet                  & 16.18${\pm}$2.04 & 36.70${\pm}$7.41 & 13.63${\pm}$1.89 & 17.90${\pm}$0.56 & 27.97${\pm}$4.47 & 22.48${\pm}$1.82  \\
GameFormer             & 11.50${\pm}$3.22 & 5.66${\pm}$2.54  & 18.82${\pm}$1.14 & 23.99${\pm}$0.83 & 25.81${\pm}$6.24 & 17.16${\pm}$1.52  \\
MAPPO                  & 5.46${\pm}$0.58  & 6.10${\pm}$0.67  & \cellcolor{green!25}7.41${\pm}$0.65  & 7.12${\pm}$0.49 & 5.38${\pm}$0.59  & 6.29${\pm}$0.27   \\
Donut                  & 15.24${\pm}$2.04 & 36.01${\pm}$6.98 & 10.80${\pm}$1.81 & 16.04${\pm}$0.52 & 24.18${\pm}$4.46 & 20.45${\pm}$1.75 \\
Self-Play RL           & \cellcolor{blue!25}2.86${\pm}$1.76 & \cellcolor{blue!25}1.54${\pm}$1.38& \cellcolor{blue!25}3.52${\pm}$1.25& \cellcolor{blue!25}2.73${\pm}$0.76& \cellcolor{green!25}3.07${\pm}$2.45  & \cellcolor{blue!25}2.74${\pm}$0.73 \\
RS-Pluto               & 7.88${\pm}$4.20 & 31.17${\pm}$4.41& 14.59${\pm}$1.81& 24.11${\pm}$5.23& 23.85${\pm}$1.99  & 20.32${\pm}$1.69 \\
TrafficGamer           & \cellcolor{green!25}3.89${\pm}$0.34 & \cellcolor{green!25}4.06${\pm}$0.55 & 8.56${\pm}$0.37 & \cellcolor{green!25}6.29${\pm}$0.50 & \cellcolor{blue!25}1.21${\pm}$0.61 & \cellcolor{green!25}4.80${\pm}$0.22 \\
\hline
\multicolumn{7}{c}{\textbf{Scenario 5}} \\ \hline
QCNet                  & 7.63${\pm}$1.35  & 15.90${\pm}$3.66 & 13.86${\pm}$0.38 & 7.36${\pm}$0.66  & 11.97${\pm}$2.57 & 11.34${\pm}$0.95  \\
GameFormer             & 17.92${\pm}$4.02 & 51.06${\pm}$13.56& 23.32${\pm}$3.55 & 30.82${\pm}$8.50 & 7.57${\pm}$1.62  & 26.14${\pm}$3.39 \\
MAPPO                  & \cellcolor{green!25}4.25${\pm}$0.60  & \cellcolor{blue!25}4.04${\pm}$0.50  & 8.21${\pm}$0.78  & 6.62${\pm}$1.01  & 8.18${\pm}$0.91  & 6.26${\pm}$0.35   \\
Donut                  & 4.90${\pm}$0.93  & 5.58${\pm}$3.63  & 7.77${\pm}$0.25  & \cellcolor{blue!25}2.59${\pm}$0.46  & 9.86${\pm}$2.54  & \cellcolor{green!25}6.14${\pm}$0.95   \\
Self-Play RL           & 14.11${\pm}$4.40 & 9.97${\pm}$3.70& \cellcolor{blue!25}3.11${\pm}$1.75& 13.24${\pm}$3.30& 2.96${\pm}$1.94  & 8.68${\pm}$1.43 \\
RS-Pluto               & 17.83${\pm}$1.77 & 11.20${\pm}$0.98& 12.60${\pm}$0.23& 17.97${\pm}$1.86& \cellcolor{green!25}4.08${\pm}$0.33  & 12.74${\pm}$0.56 \\
TrafficGamer           & \cellcolor{blue!25}4.14${\pm}$0.40 & \cellcolor{green!25}4.15${\pm}$0.79 & \cellcolor{green!25}5.38${\pm}$0.31 & \cellcolor{green!25}3.21${\pm}$0.69  & \cellcolor{blue!25}1.78${\pm}$0.30 & \cellcolor{blue!25}3.73${\pm}$0.24 \\
\hline
\multicolumn{7}{c}{\textbf{Scenario 6}} \\ \hline
QCNet                  & 33.58${\pm}$4.52 & 28.68${\pm}$10.28& 77.72${\pm}$15.58& 48.69${\pm}$12.74& 11.82${\pm}$2.35 & 40.10${\pm}$4.63 \\
GameFormer             & 57.76${\pm}$13.98& 71.80${\pm}$25.95& 52.20${\pm}$16.83& 22.50${\pm}$4.72 & 17.42${\pm}$1.54 & 44.34${\pm}$6.86 \\
MAPPO                  & 6.95${\pm}$1.21  & 6.50${\pm}$0.79  & 14.41${\pm}$0.82 & \cellcolor{green!25}5.44${\pm}$0.82  & \cellcolor{green!25}3.03${\pm}$0.61  & 7.27${\pm}$0.39   \\
Donut                  & 27.00${\pm}$4.50 & 23.96${\pm}$10.37& 69.77${\pm}$14.45& 46.42${\pm}$12.83& 6.19${\pm}$1.63  & 34.67${\pm}$4.49 \\
Self-Play RL           & \cellcolor{green!25}1.12${\pm}$1.13 & \cellcolor{green!25}2.89${\pm}$1.41& \cellcolor{green!25}9.36${\pm}$1.28& \cellcolor{blue!25}0.57${\pm}$0.33& 6.82${\pm}$1.12  & \cellcolor{green!25}4.15${\pm}$0.50 \\
RS-Pluto               & 13.01${\pm}$1.59 & 28.15${\pm}$14.09& 49.48${\pm}$8.17& 15.17${\pm}$3.26& 6.41${\pm}$2.05  & 22.44${\pm}$3.36 \\
TrafficGamer           & \cellcolor{blue!25}0.66${\pm}$0.53 & \cellcolor{blue!25}1.90${\pm}$0.56 & \cellcolor{blue!25}3.58${\pm}$1.33 & 7.63${\pm}$0.66 & \cellcolor{blue!25}2.54${\pm}$0.34 & \cellcolor{blue!25}3.26${\pm}$0.34 \\
\hline
\end{tabular}
}
\end{table}

\subsection{Fidelity Validation of Generated Motion Trajectories} 

\begin{figure}[htbp]
\centering
\centerline{\includegraphics[width=\columnwidth]{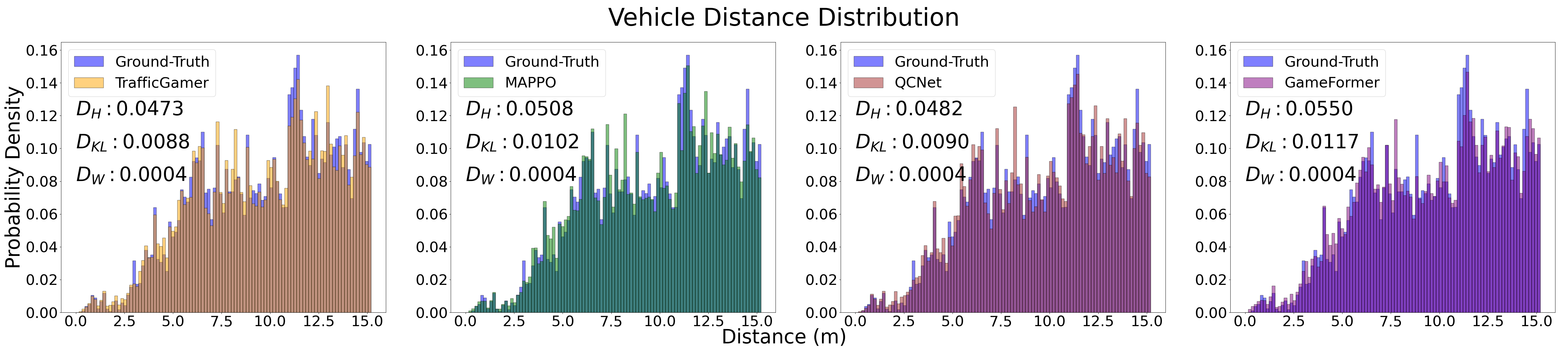}}
\centerline{\includegraphics[width=\columnwidth]{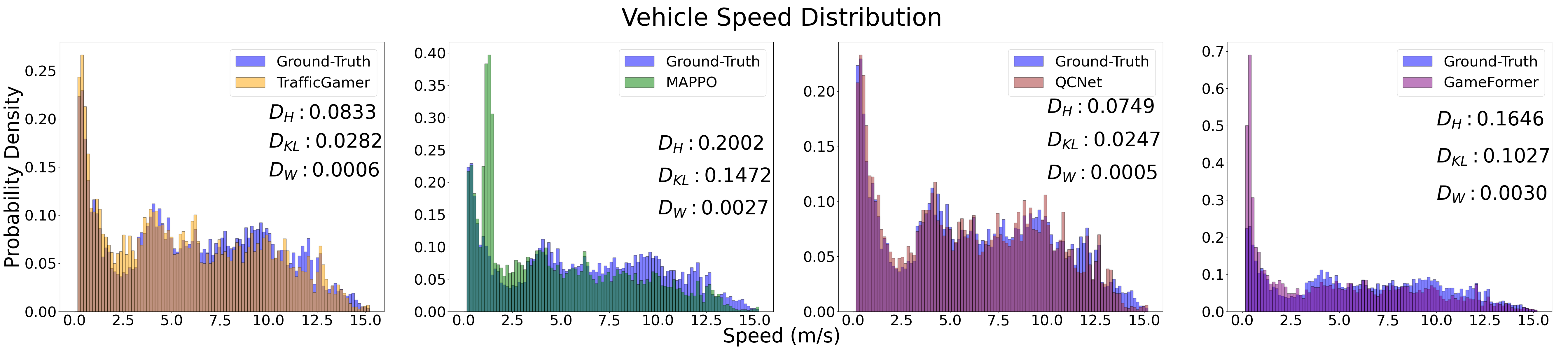}}
\centerline{\includegraphics[width=\columnwidth]{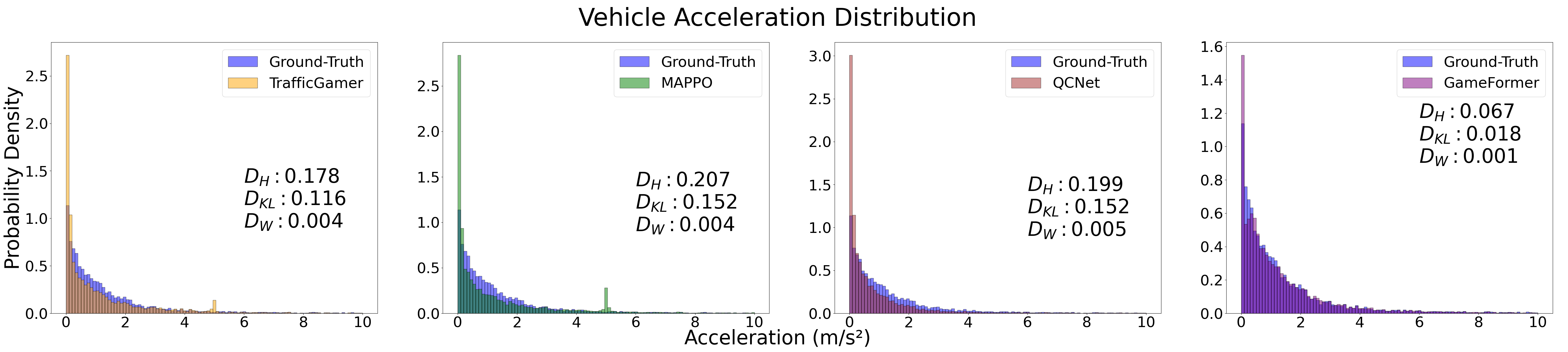}}
\centerline{\includegraphics[width=\columnwidth]{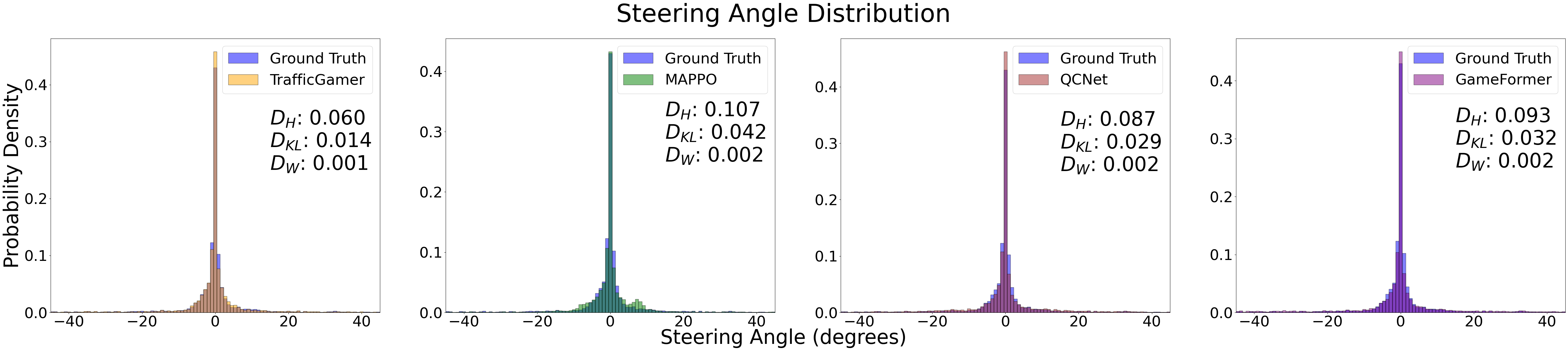}}
\caption{{\bf Statistical realism of driving behavior.} From top to bottom are vehicle distance, velocity, acceleration, and steering angle distributions.}
\label{fig: fidelity}
\end{figure}

A crucial prerequisite for a safety-critical traffic simulation is that it aligns with realistic traffic scenarios. 
We characterize this alignment with distributional fidelity, which quantifies the divergence between realistic traffic distributions and those simulated by our \md.
Among the spatial and temporal traffic features, vehicle speed, acceleration, steering angle and inter-vehicle distance are the most commonly used features for examining the performance of AV simulators\cite{yan2023learning}. We adapt $f$-divergence to qualify how well the simulated distributions match the realistic ones from the large-scale traffic datasets (Argoverse 2). The results of WOMD can be checked in the Supplementary Material. In this experiment, we assess fidelity by quantifying the distributional divergence using several fidelity metrics (Fig. \ref{fig: fidelity}). We find the simulated scenarios generated by our \md\; can properly imitate the real-world distribution of instantaneous vehicle velocity, distance, acceleration, and steering angle. 
Specifically, for methods based on {\it multi-agent fine-tuning} (\md\; and MAPPO), we observe a smaller divergence between realistic traffic data and the simulated traffic features. This is because the MMD objective in our \md\; (Objective~\ref{obj:marl-kl-mmd}) explicitly minimizes the divergence from the observed policy during optimization. The reduction in divergence results in a more accurate replication of vehicle dynamic distributions, indicating that the learned multi-agent policies are more closely aligned with actual human driving behaviors. Moreover, the performance of \md\; is comparable to that of imitation learning-based methods (QCNet and GameFormer) in terms of ensuring the fidelity of realistic driving behaviors.

\begin{figure*}[htbp]

    \begin{minipage}[t]{0.33\textwidth}
           \centerline{\includegraphics[width=\columnwidth]{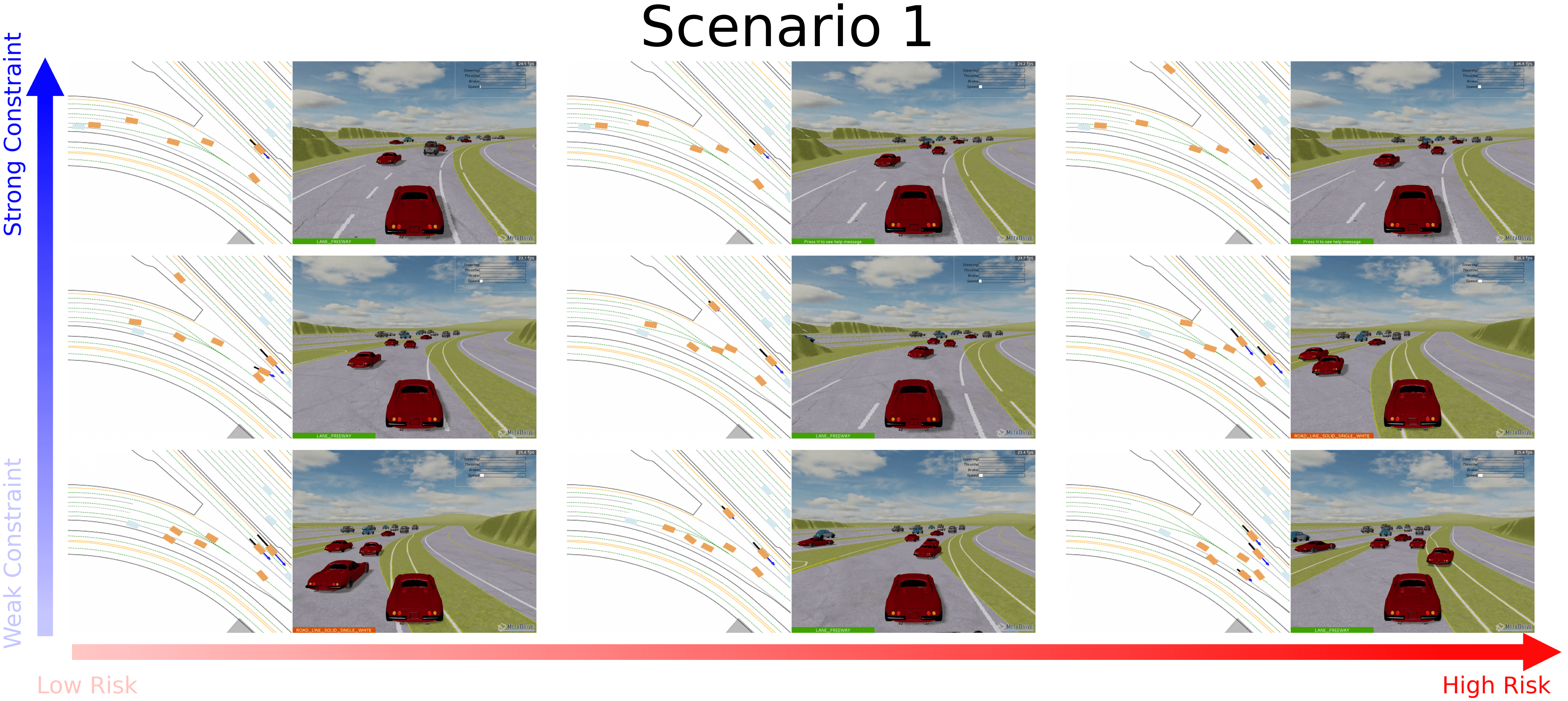}}
    \end{minipage}
    \begin{minipage}[t]{0.33\textwidth}
        \centerline{\includegraphics[width=\columnwidth]{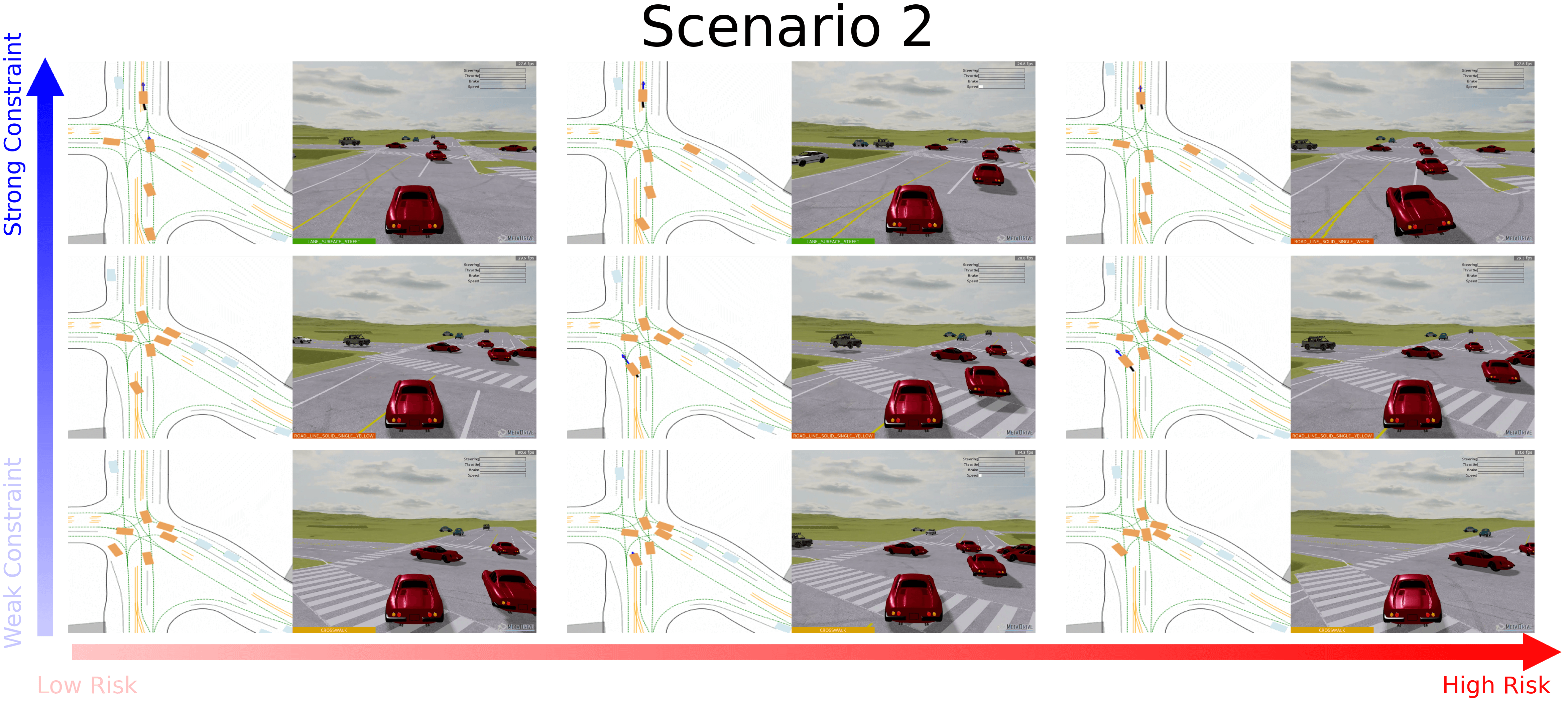}}
    \end{minipage}
    \begin{minipage}[t]{0.33\textwidth}
        \centerline{\includegraphics[width=\columnwidth]{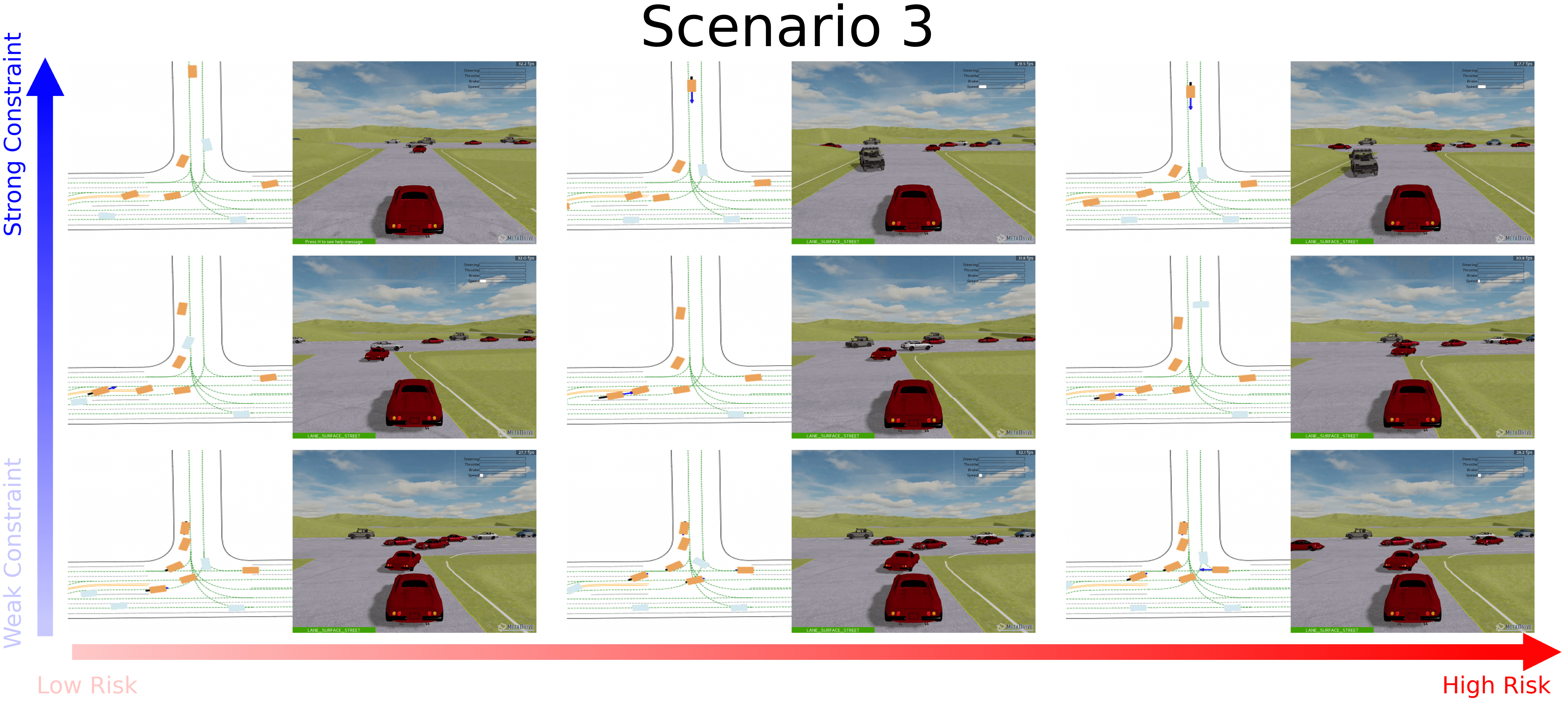}}
    \end{minipage}

    \begin{minipage}[t]{0.33\textwidth}
        \centerline{\includegraphics[width=\columnwidth]{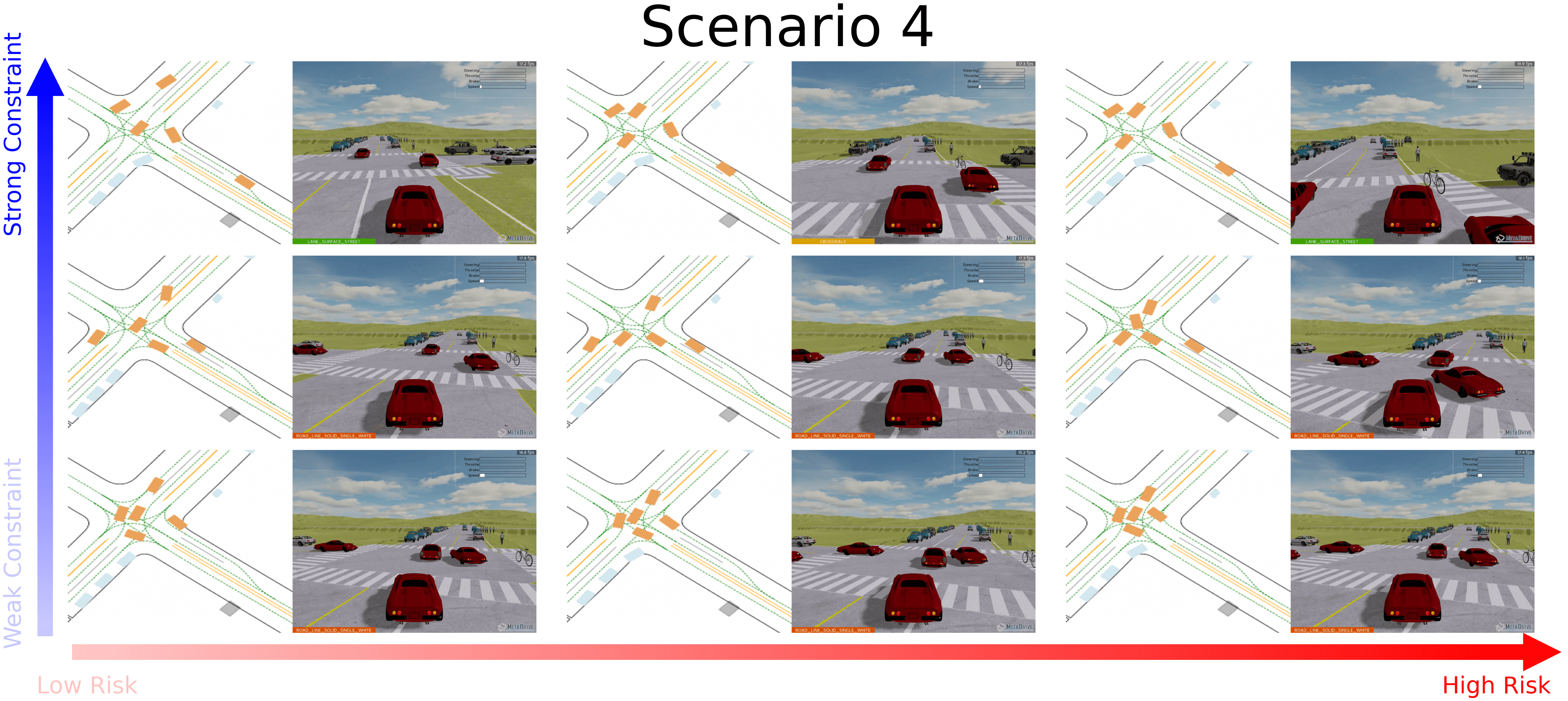}}
    \end{minipage}
    \begin{minipage}[t]{0.33\textwidth}
    \centerline{\includegraphics[width=\columnwidth]{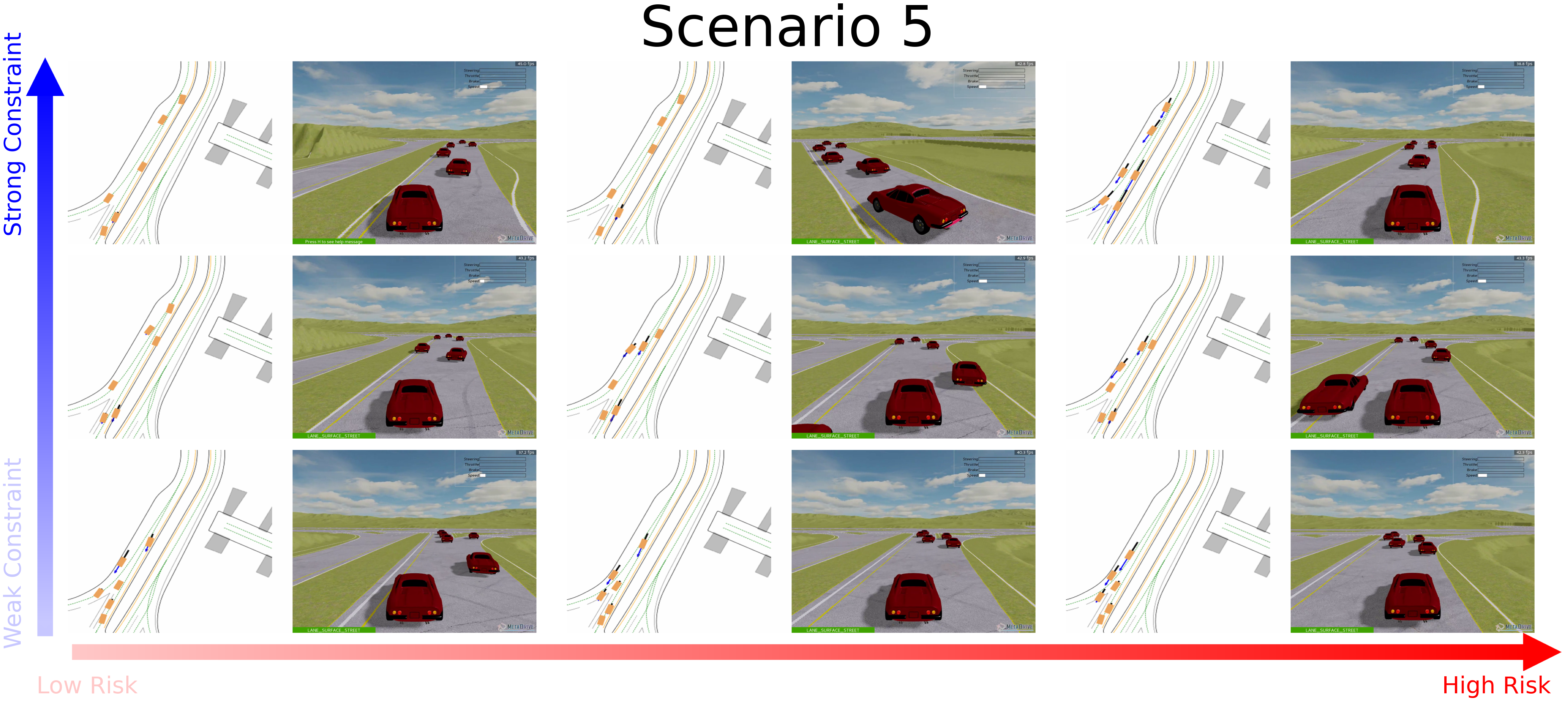}}
    \end{minipage}
    \begin{minipage}[t]{0.33\textwidth}
     \centerline{\includegraphics[width=\columnwidth]{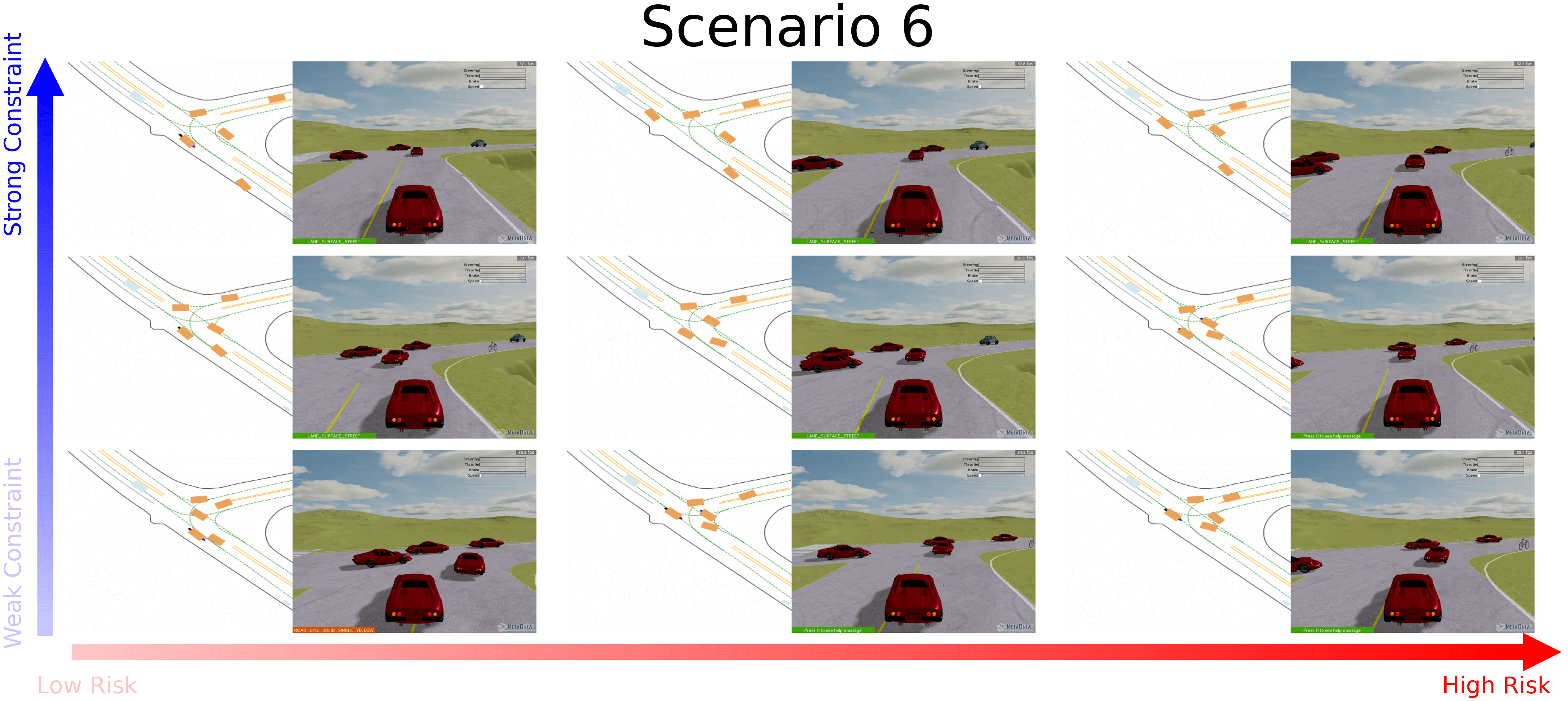}}
    \end{minipage}
    \caption{{\bf Visualization of flexible generated trajectories (WOMD).} In the 2D scenes, orange cars are algorithm-controlled, gray cars are environmental vehicles, and blue arrows indicate travel direction. In the 3D scenes, red cars are algorithm-controlled, while blue cars represent environmental vehicles.}
    \label{fig: simulation2}
\end{figure*}

\subsection{Generalization Across Diverse Datasets and Models} 

To validate the generalization of TrafficGamer, we also conduct experiments on another real-world dataset WOMD\cite{Ettinger2021waymodataset}. The setting is the same as the Argoverse 2 experiment. We can similarly observe traffic scenarios with varying levels of risk and constraint in Fig. \ref{fig: simulation2}. Multiple viewing perspectives effectively capture variations in inter-vehicle competitiveness, which generates CCEs of different intensities and consequently creates traffic scenarios with varying density levels. In addition, we also extend this algorithm to a new traffic simulation engine, GPUDrive\cite{kazemkhani2025gpudrive} based on the WOMD dataset for evaluating the fidelity of our method. In this simulator, we set that each vehicle only needs to reach the vicinity of the ground truth destination, and the observation angle for each vehicle is 120 degrees. We train the policy using 10,000 scenario samples and categorize collision types as ignore, stop, and remove for evaluation. 
The Supplementary Material clarifies the differences between the configurations of these collision types and demonstrates that TrafficGamer is still able to maintain a low crash rate, a minimal off-road rate, and a high goal-achieved rate.

\section{Conclusion}
\label{sec: conclusion}
We have demonstrated that TrafficGamer can effectively generate diverse and realistic safety-critical traffic scenarios, including commonly observed congestion patterns such as roundabouts, intersections, and merges. Compared to data-driven imitation learning methods, we actively explore strategies with varying risks, rather than passively imitating; compared to adversarial methods, we focus on the interaction strategies of all participants, rather than directly identifying the adversarial participant, as the identification directly impacts the effectiveness of the generation. However, the complexity of real-world scenarios, with numerous participants and diverse behaviors, makes identification challenging. Compared to knowledge-driven approaches, our rules are endogenous and are equilibrium strategies learned through the game, rather than externally imposed rigid rules. To enable reliable and flexible scenario generation, we address three key challenges: (1) controlling scenario diversity, (2) capturing CCEs by modeling vehicle interaction dynamics, and (3) ensuring the fidelity of generated trajectories. In contrast to traditional simulation systems that rely on hand-crafted rules or trajectory imitation often lacking realism and variability, TrafficGamer offers a principled approach to constructing high-fidelity test environments for evaluating the robustness of AV control algorithms prior to real-world deployment.

One limitation of TrafficGamer is that it may be influenced by the number of vehicles participating in the equilibrium solution process within the environment. For TrafficGamer's multi-agent CCE-Solver, solving an equilibrium for the entire map is significantly more challenging than solving it for a local map. Each additional vehicle increases both the complexity of the solution and the consumption of resources. However, regarding the dimensionality issue, for the policy network and cost value network in our method, the input dimension remains fixed. For the critic network, the input dimension needs to be increased, since the critic receives the global state as input rather than the individual observations of each vehicle. To address this, we propose a promising approach where some simulators can parallelize training using a large number of environments in simulations, which helps accelerate convergence even when the number of interacting vehicles is large. Each environment can either train a single traffic scenario or employ Multi-task reinforcement learning to parallelize training across multiple scenario tasks. 


The impact of AV‘s prevalence on the framework’s performance is indeed a critical issue concerning the real-world applicability of our work. However, the generation of safety-critical scenarios fundamentally stems from complex interactions between vehicles. Whether these interactions occur between AVs, or between AVs and human-driven vehicles, the underlying logic of interaction remains analogous. Our framework is inherently independent of the number and type of agents involved in the interaction. It solves for equilibria among multiple participants in a given scenario. In the future, AV technology will undoubtedly mature further, and it will increase the reliance on testing for safety-critical scenarios, making our research particularly meaningful.


\bibliographystyle{ieeetr}
\bibliography{TAI/reference}

\begin{IEEEbiography}
[{\includegraphics[width=1in,height=1.25in,clip,keepaspectratio]{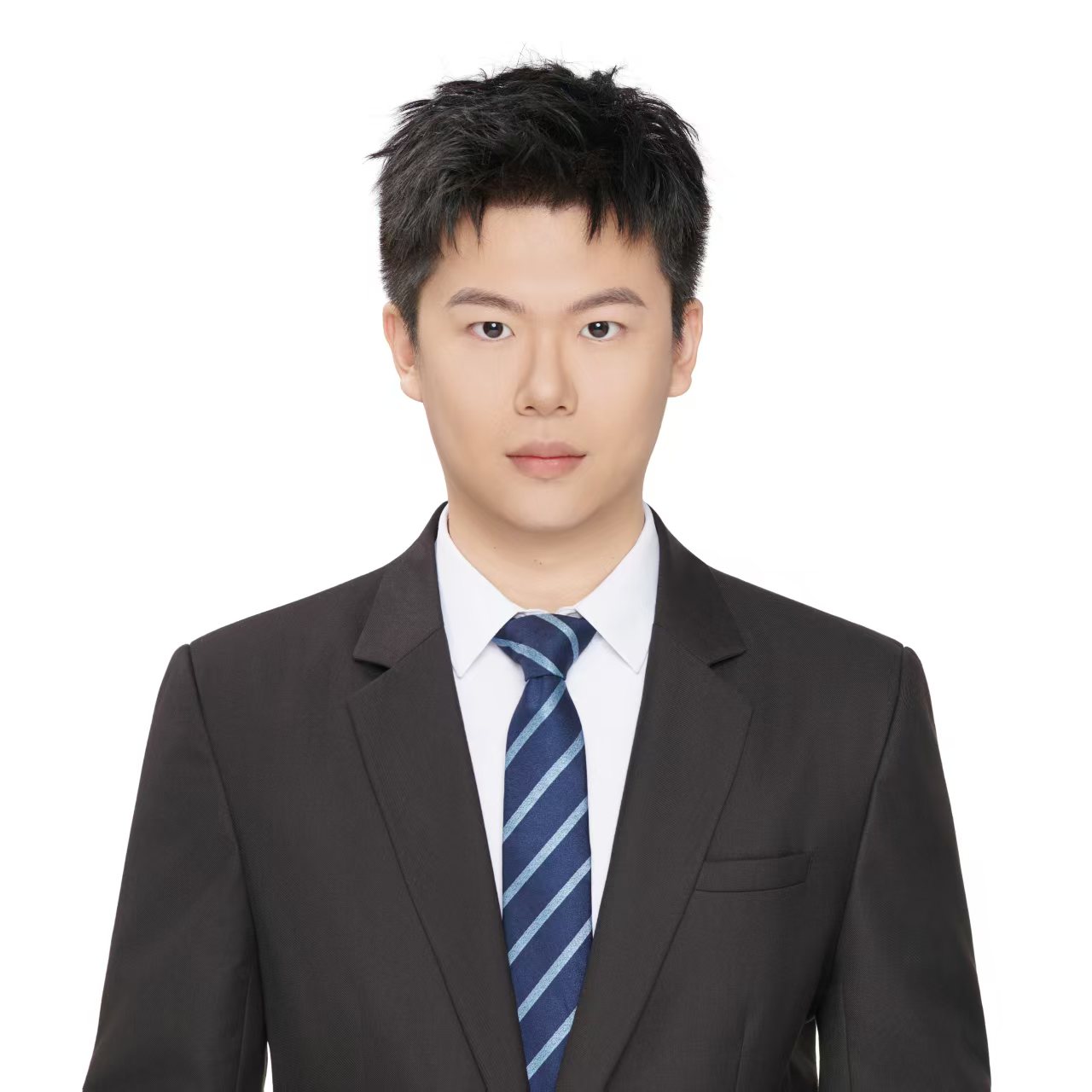}}]
{Guanren Qiao} is now a Ph.D. candidate at
School of Data Science in the Chinese University of Hong Kong, Shenzhen under the supervision of Prof. Guiliang Liu. Before that, he received the B.S. degree at Dept. of Computer Science from Beijing University of Posts and Telecommunications in 2023. His research interests
lie in robot locomotion and manipulation, reinforcement learning, and the vision-language-action model.
\end{IEEEbiography}
\begin{IEEEbiography}
[{\includegraphics[width=1in,height=1.25in,clip,keepaspectratio]{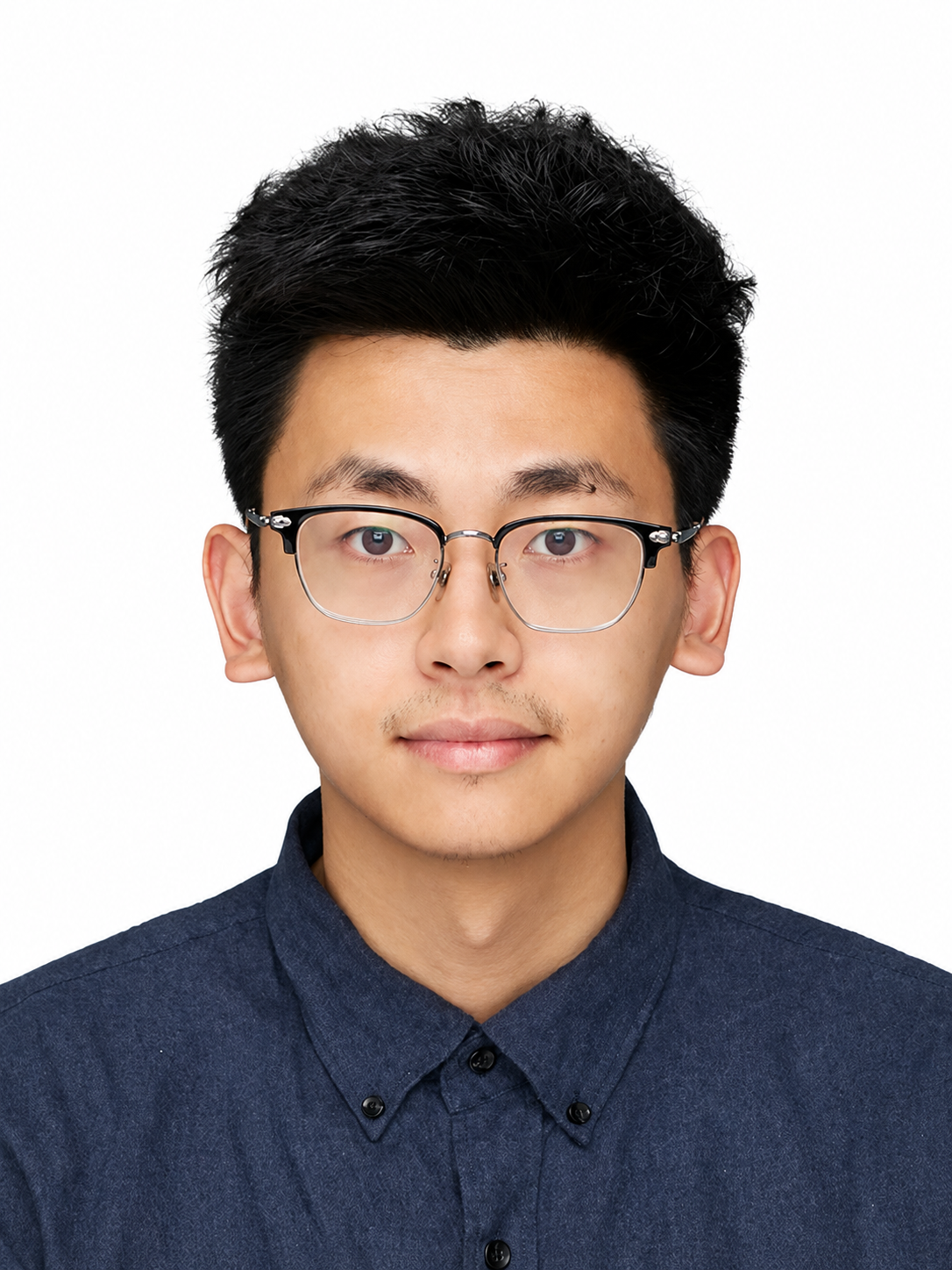}}]
{Guorui Quan} is a Ph.D. candidate at the University of Manchester, under the supervision of Prof. Manuel López-Ibáñez. Previously, he received his M.S. degree from the Institute of Statistics and Big Data at Renmin University of China. His research interests include AI-powered heuristics for industrial mathematical optimization and automated heuristic design using LLMs.
\end{IEEEbiography}
\begin{IEEEbiography}
[{\includegraphics[width=1in,height=1.25in,clip,keepaspectratio]{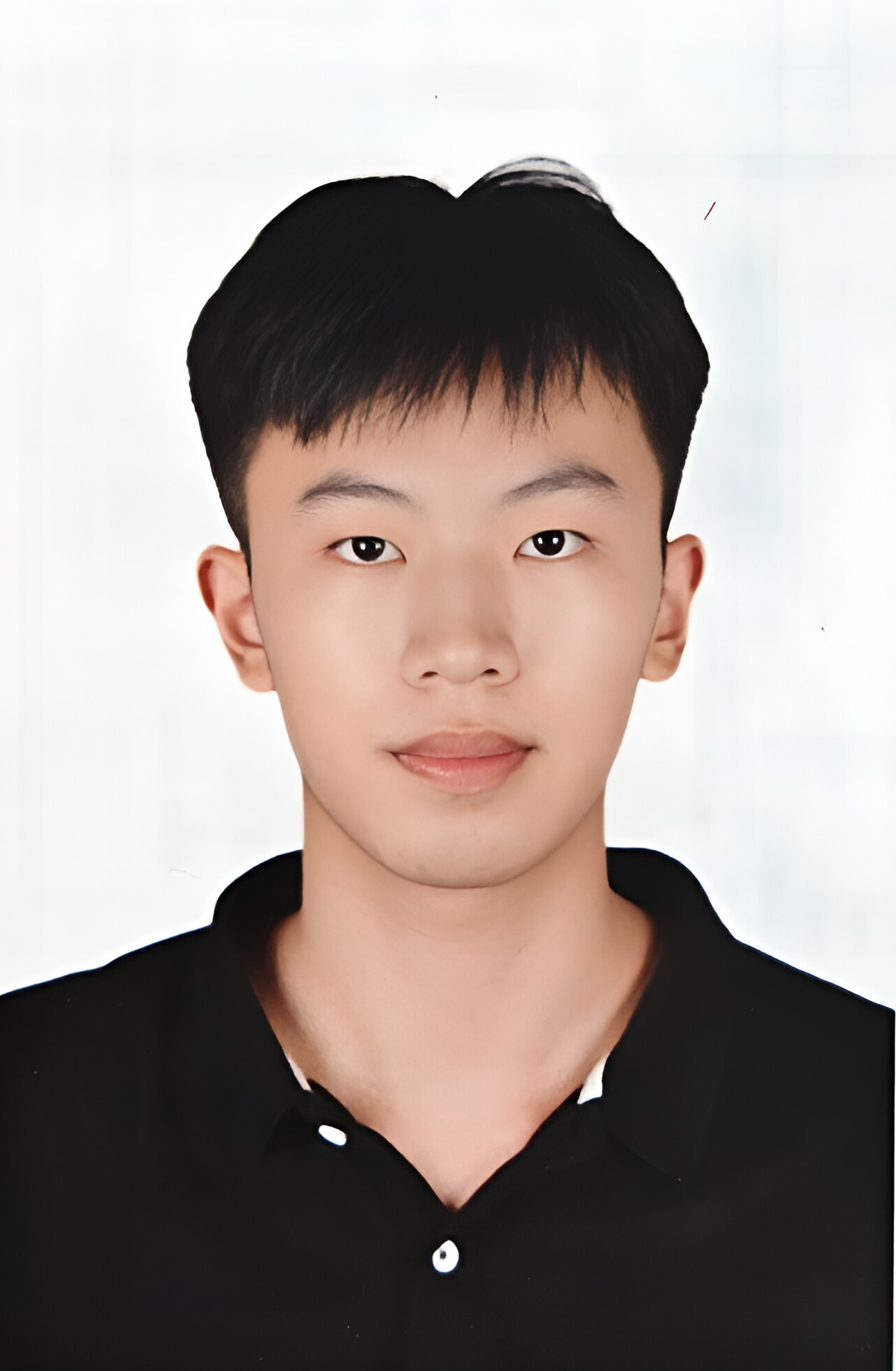}}]
{Jiawei Yu} received the B.S. degree in Computer Science and Engineering from The Chinese University of Hong Kong, Shenzhen. His research interests lie in reinforcement learning, large language models, and agents.
\end{IEEEbiography}
\begin{IEEEbiography}
[{\includegraphics[width=1in,height=1.25in,clip,keepaspectratio]{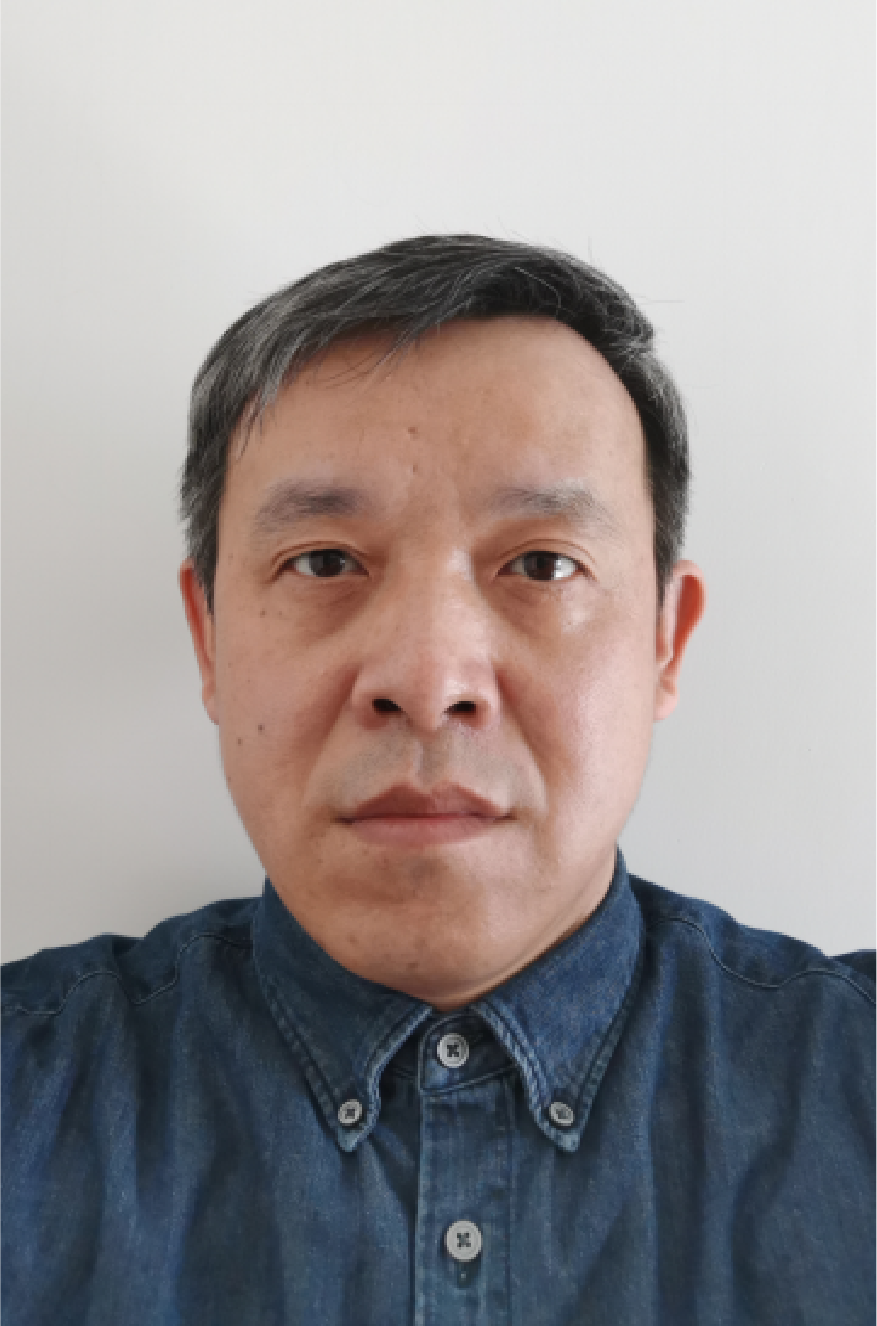}}]
{Shujun Jia}, Senior Engineer, serves as Director of the AI Innovation Center at Intelligent Road Cloud (Liaoning) Transportation Technology Co., Ltd. He earned his B.S. in Mechanical Design and Manufacturing from Northeastern University in 1997. With 20+ years in software development, algorithm R$\&$D and technical management, he has over a decade of industrial AI research. His interests cover computer vision, autonomous driving, intelligent transportation and evolvable machine learning including reinforcement and active learning.
He has partnered with CUHK(SZ) and Tsinghua University on research, co-authored a Tsinghua image processing textbook, published IEEE $\&$ arXiv papers, and holds over ten granted invention patents for visual perception, autonomous driving and transportation scenarios.
\end{IEEEbiography}
\begin{IEEEbiography}
[{\includegraphics[width=1in,height=1.25in,clip,keepaspectratio]{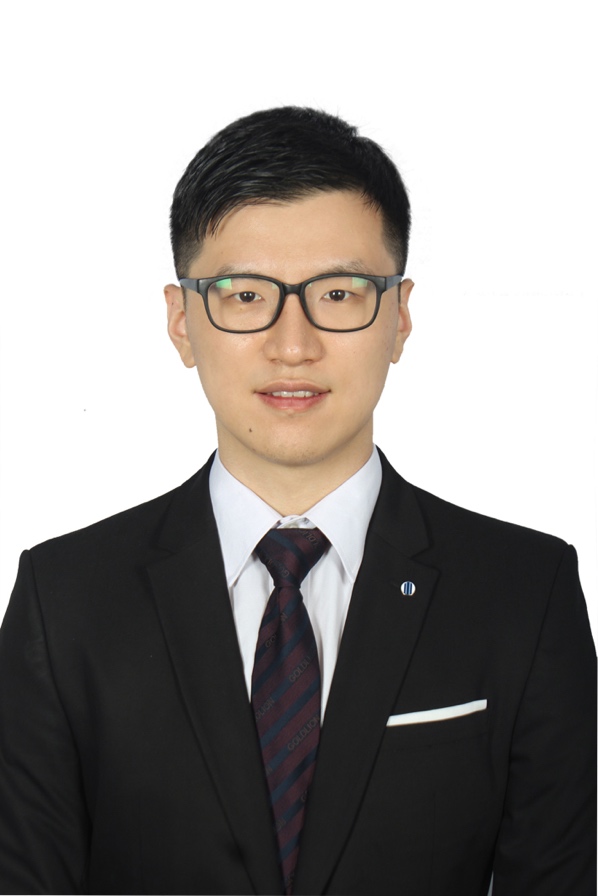}}]
{Dr. Guiliang Liu} is an Assistant Professor at the School of Data Science, The Chinese University of Hong Kong, Shenzhen. Dr. Liu’s research centers on reinforcement learning and embodied intelligent decision-making. He has pioneered the development of data engines for embodied intelligence, facilitating the generation and deployment of robotic manipulation skills through sim-to-real generalization. He has also introduced inverse constrained reinforcement learning models to improve the safety and stability of reinforcement learning control systems. In addition to his academic roles, Dr. Liu serves as the Chief Scientist for Reinforcement Learning at DexForce, the Director of the Embodied Decision Making (EDeM) lab, as well as a Research Fellow at the Shenzhen Loop Area Institute. Dr. Liu is currently an Area Chair for NeurIPS ,ICLR, CoRL and has been recognized with several prestigious accolades.
\end{IEEEbiography}
\end{document}